\documentclass[10pt,journal,compsoc]{IEEEtran}

\ifCLASSOPTIONcompsoc
  \usepackage[nocompress]{cite}
\else
  \usepackage{cite}
\fi

\ifCLASSINFOpdf
\else
\fi

\usepackage{float}
\usepackage{amsmath}
\usepackage{amsfonts}
\usepackage{algorithmic}
\usepackage{graphicx}
\usepackage{mathtools}
\usepackage{multirow}
\usepackage[marginal]{footmisc}
\usepackage{ntheorem}
\usepackage{url}
\usepackage{subfigure}
\usepackage{verbatim}

\newtheorem*{prf}{Proof}

\usepackage{color}

\newtheorem{theorem}{Theorem}
\newcommand{\bm}[1]{\mbox{\boldmath{$#1$}}}
\def\x{\mathbf{x}}
\def\e{\mathbf{e}}
\def\y{\mathbf{y}}

\def\pPhi{\mathbf{\Phi}}

\def\s{\mathbf{s}}
\def\h{\mathbf{h}}
\def\w{\boldsymbol{w}}

\def\E{\mathcal{E}}
\def\LL{\mathcal{L}}
\def\FF{\mathcal{F}}

\def\R{\mathbb{R}}
\def\NN{\mathbb{N}}
\def\N{\mathbb{N}}

\DeclareMathOperator{\X}{X}
\DeclareMathOperator{\Y}{Y}

\begin{document}

\title{Collaborative Uncertainty Benefits Multi-Agent Multi-Modal Trajectory Forecasting}
\author{Bohan Tang$^*$,
        Yiqi Zhong$^*$,
        Chenxin Xu,
        Wei-Tao Wu,
        Ulrich Neumann,~\IEEEmembership{Member,~IEEE,}
        Yanfeng Wang,
        Ya Zhang,~\IEEEmembership{Member,~IEEE,}
        and~Siheng Chen~\IEEEmembership{Member,~IEEE}

\thanks{$^*$ The first two authors share equal contributions.}
\IEEEcompsocitemizethanks{
\IEEEcompsocthanksitem B. Tang is with the Oxford-Man Institute and the Department of Engineering Science, University of Oxford, Oxford OX2 6ED, UK. This work was mainly done while he was an undergraduate student at Shanghai Jiao Tong University. \protect\\
E-mail: bohan.tang@eng.ox.ac.uk
\IEEEcompsocthanksitem Y. Zhong and U. Neumann are with the Computer Graphics and Immersive Technologies (CGIT) laboratory at the University of Southern California, Los Angeles, United States. \protect\\
E-mail: {yiqizhon, uneumann}@usc.edu

\IEEEcompsocthanksitem C. Xu, Y. Wang, Y. Zhang and S. Chen are with the Cooperative Medianet Innovation Center (CMIC) at
Shanghai Jiao Tong University, Shanghai, China. 
\protect\\
E-mail: {xcxwakaka, wangyanfeng, ya\_zhang, sihengc}@sjtu.edu.cn

\IEEEcompsocthanksitem Wei-Tao Wu is with School of Mechanical Engineering, Nanjing University of Science and Technology, Nanjing, China. 
\protect\\
E-mail: weitaowwtw@njust.edu.cn

\IEEEcompsocthanksitem Corresponding authors are Siheng Chen and Yanfeng Wang. \protect \\
}}


\IEEEtitleabstractindextext{
\begin{abstract}
In multi-modal multi-agent trajectory forecasting, two major challenges have not been fully tackled: 1) how to measure the uncertainty brought by the interaction module that causes correlations among the predicted trajectories of multiple agents; 2) how to rank the multiple predictions and select the optimal predicted trajectory. In order to handle aforementioned challenges, this work first proposes a novel concept, \emph{collaborative uncertainty} (CU), which models the uncertainty resulting from interaction modules. Then we build a general CU-aware regression framework with an original permutation-equivariant uncertainty estimator to do both tasks of regression and uncertainty estimation. Further, we apply the proposed framework to current SOTA multi-agent multi-modal forecasting systems as a plugin module, which enables the SOTA systems to 1) estimate the uncertainty in the multi-agent multi-modal trajectory forecasting task; 2) rank the multiple predictions and select the optimal one based on the estimated uncertainty. We conduct extensive experiments on a synthetic dataset and two public large-scale multi-agent trajectory forecasting benchmarks. Experiments show that: 1) on the synthetic dataset, the CU-aware regression framework allows the model to appropriately approximate the ground-truth Laplace distribution; 2) on the multi-agent trajectory forecasting benchmarks, the CU-aware regression framework steadily helps SOTA systems improve their performances. Specially, the proposed framework helps VectorNet improve by $262$ cm regarding the Final Displacement Error of the chosen optimal prediction on the nuScenes dataset; 3) for multi-agent multi-modal trajectory forecasting systems, prediction uncertainty is positively correlated with future stochasticity; and 4) the estimated CU values are highly related to the interactive information among agents. The proposed framework is able to guide the development of more reliable and safer forecasting systems in the future. 
\end{abstract}

}

\maketitle

\IEEEdisplaynontitleabstractindextext
\IEEEpeerreviewmaketitle

\IEEEraisesectionheading{\section{Introduction}\label{sec:Introduction}}

\IEEEPARstart{M}{ulti-agent} multi-modal trajectory forecasting is a task that aims to predict multiple future trajectories of multiple agents based on their observed trajectories and surroundings~\cite{DBLP:journals/thms/StahlDJ14,2016Supporting}. Precise trajectory forecasting provides essential information for decision-making and safety in numerous real-world applications such as self-driving cars~\cite{liang2020learning,Gao_2020_CVPR,Ye_2021_CVPR,gilles2021home}, drones~\cite{DBLP:conf/icc/XiaoZHY19}, and industrial robotics~\cite{DBLP:conf/icml/JetchevT09,DBLP:conf/aimech/RosmannO0B17}.

As deep learning rapidly advances, a number of deep-learning-based algorithms have been proposed to handle the multi-agent multi-modal trajectory forecasting task~\cite{liang2020learning,Gao_2020_CVPR,Ye_2021_CVPR,gilles2021home,Zhao_2019_CVPR,Choi_2019_ICCV,salzmann2020trajectron++,zeng2021lanercnn,li2020evolvegraph,DBLP:conf/nips/KosarajuSM0RS19}. Compared with common trajectory forecasting systems which only predict one future trajectory per agent, a multi-agent multi-modal trajectory forecasting system predicts multiple possible future trajectories for each agent to handle the stochasticity inherent in future forecasting. With state-of-the-art performances, many multi-agent multi-modal trajectory forecasting systems have been widely used in real-world applications. Nevertheless, two major challenges are underexplored in the existing works on deep-learning-based multi-modal forecasting: 1) how to measure the uncertainty over the prediction for each agent under a multi-agent setting; 2) how to rank the multiple predictions for each agent under a multi-modal setting.

Solving the first challenge of uncertainty estimation is practically crucial for the trajectory forecasting task, since the task is highly safety-critical while deep-learning-based trajectory forecasting methods do not always come with reliability. Nowadays, most researchers tend to use uncertainty analysis~\cite{Gal2016Uncertainty,NIPS2017_2650d608,zhao2020uncertainty} to assess the reliability of deep-learning-based systems. Specifically, in big data regimes (e.g., most deep learning systems that have a large number of available data), it is important to model aleatoric uncertainty, namely, the uncertainty regarding information aside from statistic models which data cannot explain~\cite{NIPS2017_2650d608}. Following this thread of thought, some explorations have been conducted for the task of trajectory forecasting. For example, on the basis of the framework proposed in~\cite{Gal2016Uncertainty}, existing works~\cite{DBLP:conf/corl/Jain0LXFSU19,DBLP:conf/cvpr/HongSP19,DBLP:conf/cvpr/LeeCVCTC17,DBLP:conf/eccv/FelsenLG18} utilize the predictive variance to estimate the uncertainty over each agent's prediction separately. By doing so, they assume that the prediction of an agent's future trajectories is independent from the predictions about other agents. However, recent SOTA methods~\cite{gu2021densetnt,salzmann2020trajectron++,girgis2021latent,liang2020learning,Gao_2020_CVPR,gilles2021gohome} have almost all mentioned that agents' future trajectories are, in many cases, non-independent because of agent-wise interactions; modeling agent-wise interactions is thus needed and would largely improve forecasting accuracy. When there is an interaction modeling module in the forecasting system, the module will let the prediction of each agent interact with each other, making predictions no longer mutually independent. In this circumstance, the independence assumption held by the previous uncertainty estimation methods will become invalid. As a result, the uncertainty measurements will also become inaccurate, which can lead to safety issues. To tackle this first challenge of uncertainty measurement in the multi-agent setting, we need a more sophisticated and robust measurement to capture the previously neglected uncertainty brought by correlated predictions.

The second challenge (i.e., how to rank multiple predictions) is related to the practical use of forecasting systems. When multiple future trajectories are predicted for an agent, it will be helpful for the system to have a ranking strategy that appropriately assigns priorities to the predictions. This ensures the efficiency of the downstream decision-making system to a great extent by letting it know which predictions should be prioritized during the decision-making procedures and require extra attention. Without a ranking strategy, the downstream system would treat all the predictions equally, which is unideal. Existing works~\cite{girgis2021autobots, gu2021densetnt,ngiam2021scene,liang2020learning,gilles2021gohome} usually adopt a classifier that is trained to assign a higher score to the prediction closer to the ground truth. This strategy, however, may encounter robustness issues during inference because 1) with much randomness in the ground truth data, it is uninterpretable for a model to learn the priorities of multiple predictions purely from the data; 2) their strategy does not consider the uncertainty of each prediction during the ranking procedure, causing safety concerns. To make the ranking procedure more interpretable and safety-oriented, an alternative strategy from us, which we will elaborate in this work, is to use the uncertainty estimated for each prediction as the guiding information when assigning the priorities.

Based on the discussions above, we realize that fully addressing the aforementioned two challenges first requires a new uncertainty estimation framework, which better handles the uncertainty analysis when the independence of prediction is not assumed. To this end, we propose a new concept, \emph{collaborative uncertainty} (CU), to estimate the uncertainty resulting from the use of interaction modules in forecasting systems. We also coin a concept, \emph{individual uncertainty} (IU), to describe the uncertainty approximated by the predictive variance of a single agent, which is the setting of previous uncertainty estimation methods. We then introduce a novel CU-aware regression framework in Section~\ref{sec:Methodology}, which describes how to measure CU and IU simultaneously in a general regression task. This framework utilizes the mean $\mu$ and covariance $\Sigma$ of a predictive distribution $p(\Y|\X)$ to estimate the prediction result and its corresponding uncertainty, where $\X$ is the input data and $\Y$ is the regression target. Furthermore, this framework contains a regression model with an original permutation-equivariant uncertainty estimator that learns the values of mean $\mu$ and covariance $\Sigma$.

\begin{figure}[tb]
    \centering
    \includegraphics[width=1.\columnwidth]{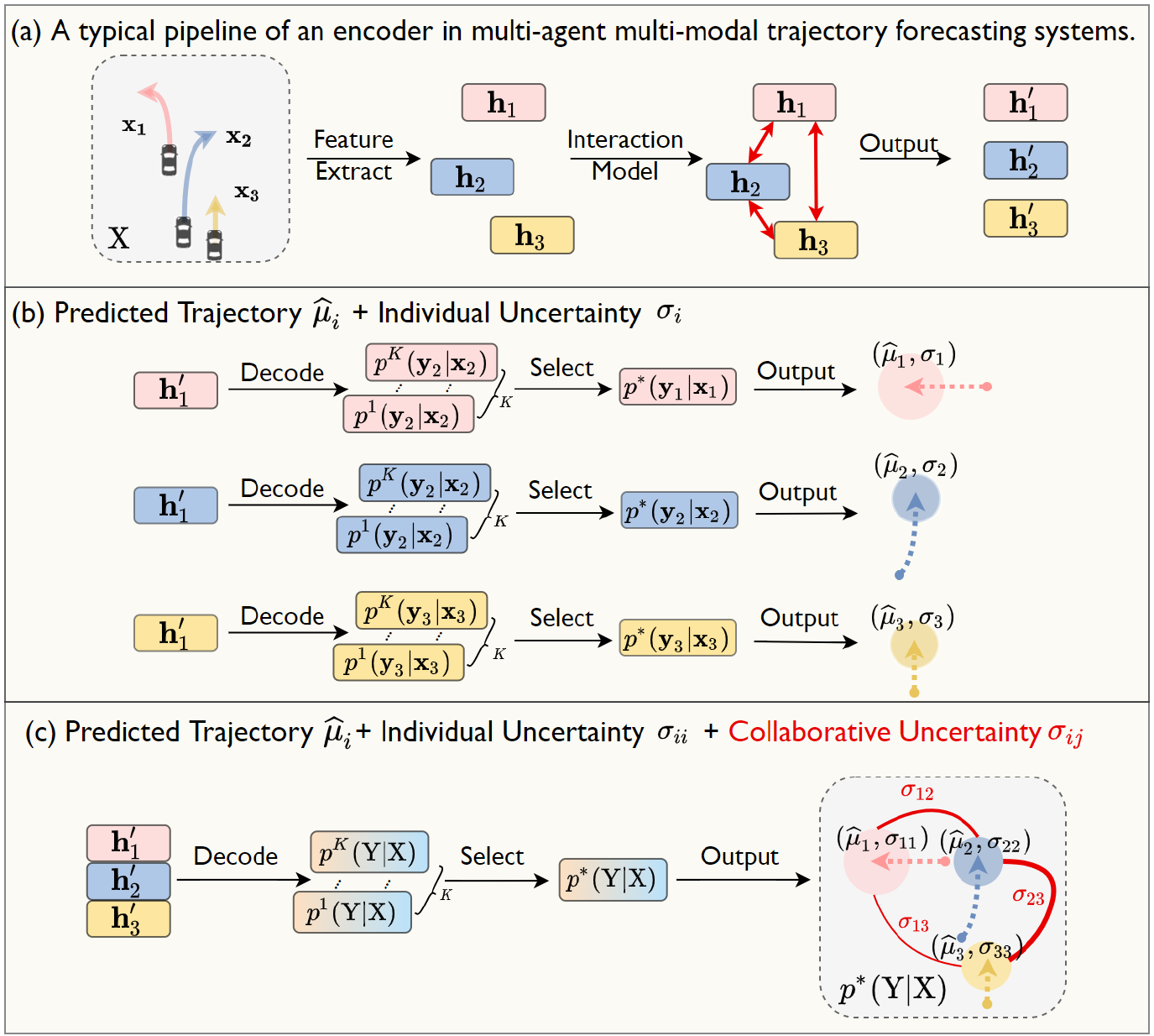}
    \caption{\textbf{Uncertainty estimation in multi-agent multi-modal trajectory forecasting systems.} (a) A typical pipeline of an encoder in multi-agent multi-modal trajectory forecasting systems. (b), (c) respectively illustrate the decoder pipeline of previous methods and our method. Previous methods output the predicted trajectory $\widehat{\mu}_{i}$ and individual uncertainty $\sigma_i$ while our method additionally outputs \textit{collaborative uncertainty} $\sigma_{ij}$.}
    \label{fig:matf_example}
\end{figure}

On the basis of our proposed CU-aware regression framework, we introduce a CU-aware multi-agent multi-modal trajectory forecasting system, which handles the two challenges of the trajectory forecasting task: uncertainty estimation and ranking strategy. The system consists a novel CU-aware forecasting module and an original CU-based selection module. The CU-aware forecasting module utilizes the regression model of the CU-aware regression framework to learn the mappings that are from input data to i) accurate prediction; ii) individual uncertainty; and iii) collaborative uncertainty. This design leads to more precise uncertainty estimation and prediction in the multi-agent multi-modal trajectory forecasting setting, which aims to solve the first challenge, uncertainty estimation. The CU-based selection module then ranks the multi-modal predictions of each agent according to the uncertainty value estimated by the CU-aware regression framework. It encourages the system to prioritize the predictions with lower uncertainty levels, which largely improves the interpretability of the ranking strategy in the multi-modal setting, addressing the second challenge. See Fig.~\ref{fig:matf_example} for an illustration of uncertainty estimation in forecasting systems. 

We conduct extensive experiments \footnote{Our code will be available at~\url{https://github.com/MediaBrain-SJTU/Collaborative-Uncertainty}} to show that: 1) the CU-aware regression framework allows the model to appropriately approximate the ground-truth multivariate Laplace distribution on the synthetic dataset; 2) adding CU estimation benefits accurate predictions; 3) future stochasticity and prediction uncertainty of the multi-agent multi-modal trajectory forecasting system are positively correlated; and 4) CU estimation yields significantly larger performance gains in forecasting systems with interaction modules (see Fig.~\ref{fig:model_type}), confirming that CU is highly related to the usage of the interaction modeling procedure.

The contributions of this work are as follows:

$\bullet$ We propose a novel concept, \textit{collaborative uncertainty} (CU), and a novel CU-aware regression framework with an original permutation-equivariant uncertainty estimator that models the uncertainty brought by agent-wise interactions.

$\bullet$ We propose a CU-aware multi-agent multi-modal trajectory forecasting system, which leverages collaborative uncertainty to address uncertainty estimation and multi-modal ranking challenges.

$\bullet$ We conduct extensive experiments to validate the CU-aware regression framework on a synthetic dataset and two large-scale real-world datasets.

$\bullet$ On the basis of our proposed CU-aware regression framework, we disclose the positive correlation between the future stochasticity and prediction uncertainty of the multi-agent multi-modal trajectory forecasting system.

A preliminary version of this work is presented in~\cite{tang2021collaborative}. In comparison, the novelty of the current work is threefold. 

$\bullet$ This work leverages the proposed CU-aware regression framework to support the multi-modal trajectory forecasting system. It innovatively designs an uncertainty-based selection module to help the system rank the multiple predicted trajectories of each agent in a more interpretable manner. Whereas, our previous work~\cite{tang2021collaborative} is only compatible with single-modal forecasting systems. 

$\bullet$ This work proposes a novel permutation-equivariant uncertainty estimator to estimate collaborative uncertainty, while our previous work~\cite{tang2021collaborative} cannot guarantee the important property of permutation-equivariance. Section~\ref{subsec:syn_data} of the current work shows that our permutation-equivariant uncertainty estimator has led to more accurate uncertainty estimation results.

$\bullet$ This work includes new experimental results that demonstrate the competitive performance of our CU-aware forecasting system in multi-agent multi-modal trajectory forecasting. Specially, the proposed CU-based framework helps VectorNet improve by $262$ cm regarding the Final Displacement Error (FDE) of the chosen optimal predicted trajectory on the nuScenes dataset. Our previous work~\cite{tang2021collaborative} only improves VectorNet by $99$ cm (FDE) on the same dataset.

The rest of this paper is structured as follows. In Section~\ref{sec:Related_Works}, we introduce previous works in the field of aleatoric uncertainty estimation and multi-agent multi-modal trajectory forecasting. In Section~\ref{sec:problem_formulation}, we formulate the problem of multi-agent multi-modal trajectory forecasting with uncertainty estimation and demonstrate the necessity of modeling collaborative uncertainty for the task. In Section~\ref{sec:Methodology}, we propose a novel CU-aware regression framework with an original permutation-equivariant uncertainty estimator and show a special case of it based on the multivariate Laplace distribution. In Section~\ref{sec:Example}, we apply our proposed CU-aware regression framework to the multi-agent multi-modal trajectory forecasting system to enable such system to better handle both tasks of uncertainty estimation and multi-modal ranking, on the basis of the special case shown in Section~\ref{sec:Methodology}. In Section~\ref{sec:Experiments}, we conduct experiments on a synthetic dataset and two real-world datasets to evaluate the effectiveness of the CU-aware regression framework in distribution estimation and trajectory forecasting. In Section~\ref{sec:discussion}, we discuss the causes of both uncertainty and collaborative uncertainty in the task of trajectory forecasting. Finally, we conclude this paper in Section~\ref{sec:Conclusions}.
\section{Related Works}\label{sec:Related_Works}
\textbf{Aleatoric uncertainty estimation in deep learning.}  
Recent efforts are rising as to improve the measurement of aleatoric uncertainty in deep learning models. One seminal work is \cite{NIPS2017_2650d608}. It proposes a unified Bayesian deep learning framework to explicitly represent aleatoric uncertainty using predictive variance for generic regression and classification tasks. Many existing works~\cite{DBLP:conf/cvpr/KendallGC18,ayhan2018test,pmlr-v80-depeweg18a,feng2018towards,wang2019aleatoric,pmlr-v119-kong20b} follow this idea and formulate uncertainty as learned loss attenuation. For example, to make predictive-variance-based aleatoric uncertainty measurements more efficient, \cite{ayhan2018test} adds data augmentation during the test time. But, these works only pay attention to individual uncertainty. 
 
Other recent works attend to the uncertainty measurement for correlated predictive distributions. For example,~\cite{8578672} and \cite{DBLP:conf/nips/MonteiroFCPMKWG20} measure \textit{spatially correlated uncertainty} in a generative model respectively for image reconstruction and pixel-wise classification, and~\cite{hafner2021mastering} captures \textit{joint uncertainty} as discrete variables in the field of reinforcement learning. Despite these three works, our work is the first to conceptualize and measure \textit{collaborative uncertainty} in the multi-agent trajectory forecasting task. To the best of our knowledge, there are only two papers~\cite{DBLP:conf/cvpr/GundavarapuSMSJ19,girgis2021autobots} that are close to our track. \cite{DBLP:conf/cvpr/GundavarapuSMSJ19} and \cite{girgis2021autobots} model the joint uncertainty in the pose estimation task and multi-agent trajectory forecasting task respectively. However, they present several limitations: 1) they did not provide a permutation-equivariant joint uncertainty estimator; 2) they did not provide a theoretical conceptualization or definition of the uncertainty due to correlated predictive distributions; and 3) they did not analyze the causes of such uncertainty. These limitations are essential problems to tackle. Thus, in this work, we not only formulate a general framework with a permutation-equivariant uncertainty estimator, but also theoretically conceptualize \emph{collaborative uncertainty} and analyze its causes.

\textbf{Multi-agent multi-modal trajectory forecasting.}
The nature of the trajectory forecasting task is that there are often more than one plausible future trajectory. Recently, multi-modal forecasting has become the dominant research setting in the trajectory forecasting research community~\cite{MemoNet_2022_CVPR,xu2022GroupNet,bae2022npsn,9779572,gu2022stochastic,liang2020learning,zeng2021lanercnn,salzmann2020trajectron++,deo2022multimodal,chai2019multipath,dendorfer2021mg,zhou2022hivt}. This setting requires models to: i) take the observed trajectories from multiple agents and their surrounding environment (e.g., HD maps) as the inputs, and outputs the multiple possible future trajectories predicted for each agent; ii) select the optimal prediction from the predicted trajectories. To tackle the aforementioned requirements, a typical deep-learning-based multi-agent multi-modal trajectory forecasting system usually consists of a regression module that predicts future trajectories and a selection module that ranks the predictions and selects the optimal future trajectory.

In most state-of-the-art (SOTA) forecasting systems, the regression module is designed based on the encoder-decoder architecture. In the encoding process, like many other sequence prediction tasks, the model used to adopt a recurrent architecture to process the inputs~\cite{DBLP:conf/cvpr/AlahiGRRLS16,DBLP:conf/cvpr/HasanSTBGC18,DBLP:conf/cvpr/Liang0NH019}. Later, however, the graph neural networks has become a more common approach as they can significantly assist trajectory forecasting by capturing the interactions among agents~\cite{MemoNet_2022_CVPR,xu2022GroupNet,bae2022npsn,9779572,gu2022stochastic,liang2020learning,https://doi.org/10.48550/arxiv.2102.09117,Ye_2021_CVPR,Yuan_2021_ICCV,hu2020collaborative,https://doi.org/10.48550/arxiv.2202.08408,li2020evolvegraph,DBLP:conf/nips/KosarajuSM0RS19,https://doi.org/10.48550/arxiv.2107.00894}. For the decoding phase, multi-modal forecasting methods usually use multi-layer perceptrons (MLPs) to decode the hidden features~\cite{yagi2018future,robicquet2016learning,lee2017desire,Gao_2020_CVPR}. Many of these methods choose to adopt multiple MLP-based decoders during the decoding phase~\cite{liang2020learning,zeng2021lanercnn,ye2021tpcn}. Each decoder individually predicts one possible trajectory. To ensure the prediction diversity, instead of optimizing all the decoders, those methods tend to only optimize the one that is closest to the ground truth. 

As for the selection module, most SOTA systems are equipped with an MLP-based selector, which assigns the highest confidence score for the prediction that is the closest to the ground truth, to generate a confidence score for each prediction.~\cite{liang2020learning,zeng2021lanercnn,ye2021tpcn,Narayanan_2021_CVPR,Liu_2021_CVPR,Kim_2021_CVPR,song2021learning,gilles2021gohome,gu2021densetnt}.

For safety reasons, it is necessary to report the uncertainty of each predicted trajectory. Works to date about uncertainty measurements ~\cite{DBLP:conf/corl/Jain0LXFSU19,DBLP:conf/cvpr/HongSP19,DBLP:conf/cvpr/LeeCVCTC17,DBLP:conf/eccv/FelsenLG18} have appropriately modeled the interaction among multi-agent trajectories for boosting performances, but they overlook the uncertainty resulting from the correlations in predicted trajectories, and their selection modules do not take the uncertainty of predictions into consideration. We seek to fill these gaps by introducing \emph{collaborative uncertainty} (CU) and designing a CU-based selection module.

\section{Problem Formulation}

\label{sec:problem_formulation}
Consider $m$ agents in a data sample, and let $\X=[\x_{1},\x_{2},...,\x_{m}]^{T}\in\R^{2T_{-}\times m}$, $\Y=[\y_{1},\y_{2},...,\y_{m}]^{T}\in\R^{2T_{+}\times m}$ be the past observed and the future trajectories of all agents, where $\x_{i}\in\R^{2T_{-}}$ and $\y_{i}\in\R^{2T_{+}}$ are the past observed and the future trajectories of the $i$-th agent. Each $\x_i/\y_i$ consists of two-dimensional coordinates at different timestamps of $T_{-}/T_{+}$. A multi-modal trajectory forecasting system usually consists of two modules: the forecasting module and the selection module. The forecasting module is used to model the predictive distribution $p(\Y|\X)$, whose mean $\mu \in \R^{K\times 2T_{+} \times m}$ is the predicted trajectories and covariance $\Sigma\in \R^{K\times T_{+} \times m\times m}$ is the corresponding uncertainty. Here $K$ is the number of prediction modals, element $\mu_{k,t,i}$ of $\mu$ is the predicted geometric location in the $k$-th prediction modal of the $i$-th agent at timestamp $t$, and $\Sigma_{k}$ contains the individual and collaborative uncertainty of agents in the $k$-th prediction modal. More details of $\Sigma$ will be specified in Subsection~\ref{subsec:CUUEF}. Moreover, the selection module aims at selecting the optimal predicted trajectory $\mu^{*}\in\R^{2T_{+} \times m}$ from $K$ predicted trajectories $\mu$.

\begin{figure}[t]
    \centering
    \includegraphics[width=1.\columnwidth]{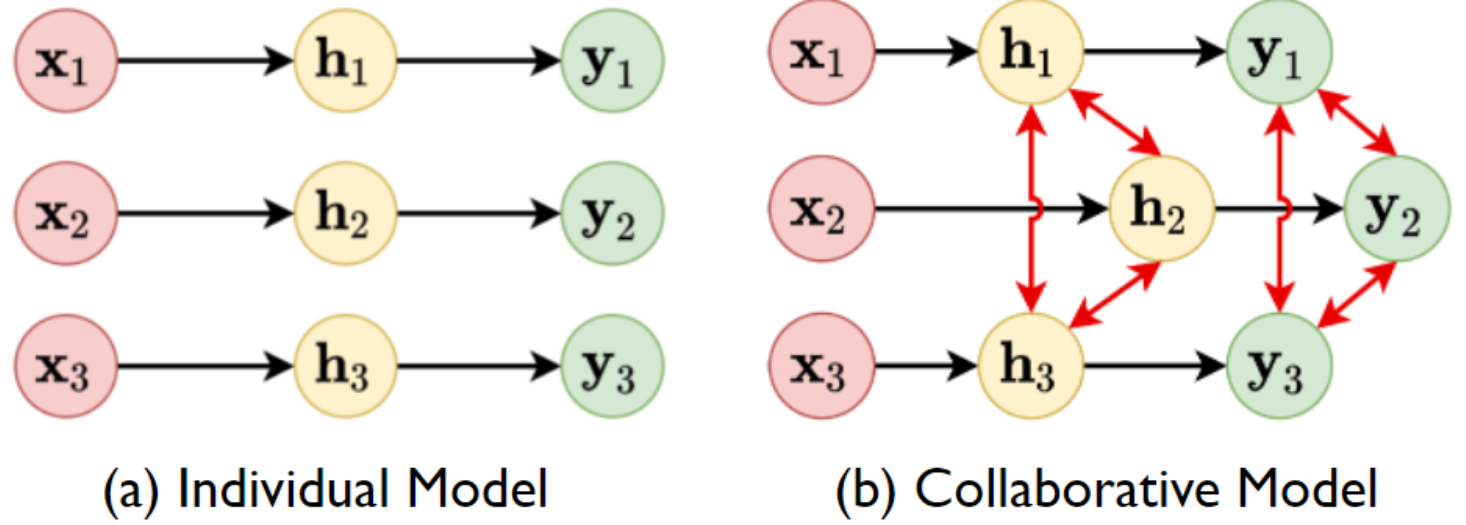}
    \caption{\textbf{Graphical models for deep learning networks in a three-agent trajectory forecasting setting}: (a) represents the model that predicts the trajectory of each agent independently; (b) shows the model that explicitly captures the interactions among multiple agents (e.g., the model containing the graph-message-passing process). $\x_{i}$ is the observed trajectory of the $i$-th agent; $\h_{i}$ and $\y_{i}$ are its corresponding hidden feature and future trajectory respectively.}
    \label{fig:model_type}
\end{figure}

In the field of uncertainty estimation, previous works such as~\cite{Gal2016Uncertainty,NIPS2017_2650d608,pmlr-v48-gal16} use the individual distribution to approximate $p(\Y|\X)$. The assumption behind this approach is that $p(\y_{i}|\x_{i})$ is independent for every $i\in\{1,2,3,...,m\}$. Mathematically, they set the covariance $\Sigma$ as a diagonal matrix. This assumption is valid for the regression task that uses the model shown in Fig.~\ref{fig:model_type} (a). We refer to the uncertainty under the independence assumption as \emph{individual uncertainty} in this paper. However, most of the cutting-edge multi-agent multi-modal trajectory regression models, such as~\cite{liang2020learning,Gao_2020_CVPR,Ye_2021_CVPR,gilles2021home,Zhao_2019_CVPR,Choi_2019_ICCV,salzmann2020trajectron++,zeng2021lanercnn,li2020evolvegraph,DBLP:conf/nips/KosarajuSM0RS19}, adopt the collaborative model to model the interactions among multiple agents (see Fig.~\ref{fig:model_type} (b) for an illustration of the collaborative model). In the collaborative model, $\y_{i}$ is no longer solely dependent on $\x_{i}$, but also on other agents $\x_{j}$ where $j\neq i$, which turns $p(\Y|\X)$ from the individual distribution into the joint distribution of multiple agents and results in a new type of uncertainty. We call this type of uncertainty brought by the interactions as \textit{collaborative uncertainty} (CU).

\begin{figure*}[t!]
\begin{center}
\centerline{\includegraphics[width = 2.\columnwidth]{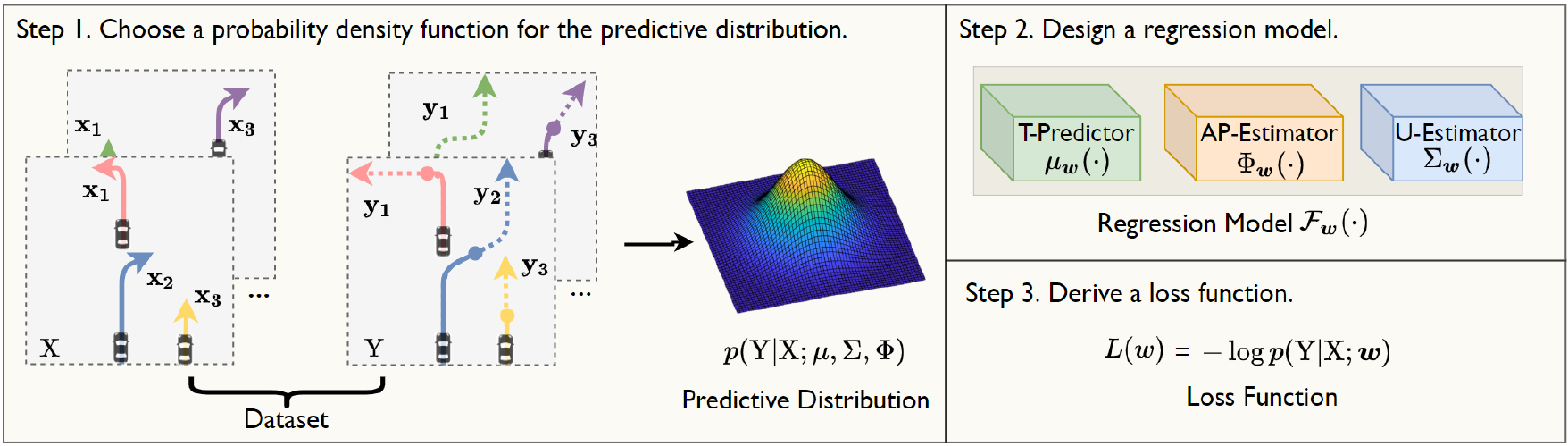}}
\caption{\textbf{Probabilistic formulation of the CU-aware regression framework.} We first choose a proper probability density function for the predictive distribution based on the given dataset. Then we design a regression model $\FF_{\w}(\cdot)$ to estimate the parameters of the chosen probability density function. Finally, on the basis of the chosen probability density function, we derive a loss function for the regression model $\FF_{\w}(\cdot)$.}
\label{fig:three_step}
\end{center}
\end{figure*}

In the next section, to delineate the complete landscape of uncertainty in multi-agent multi-modal trajectory forecasting, we introduce a unified CU-aware regression framework with an original permutation-equivariant uncertainty estimator for estimating both individual and collaborative uncertainty. Furthermore, we demonstrate a special case of it: Laplace CU-aware regression, which is used in the multi-agent multi-modal trajectory forecasting system.

\section{Collaborative Uncertainty} \label{sec:Methodology}
In this section, we first introduce a novel CU-aware regression framework with a probabilistic formulation and an original permutation-equivariant uncertainty estimator. Then we demonstrate a special case of our framework: Laplace CU-aware regression. It specifically models uncertainty when data obey the multivariate Laplace distribution, which is a widely held assumption in the multi-agent trajectory forecasting task~\cite{liang2020learning,Gao_2020_CVPR,DBLP:conf/nips/KosarajuSM0RS19,tang2021collaborative}.

\subsection{CU-Aware Regression Framework}
\label{subsec:CUUEF}

The CU-aware regression framework consists of two key components: a probabilistic formulation and a permutation-equivariant uncertainty estimator. The probabilistic formulation leads to a regression model and a loss function, which are used to estimate uncertainty in the regression task. The permutation-equivariant uncertainty estimator enables our framework to generate uncertainty permutation-equivariant with the input data.

\subsubsection{Probabilistic Formulation}\label{subsec:GFofCU}
Contrary to prior works in uncertainty estimation, to model collaborative uncertainty, we abandon the independence assumption held by previous works~\cite{Gal2016Uncertainty,NIPS2017_2650d608,pmlr-v48-gal16}, and set $p(\Y|\X)$ as a joint multivariate distribution with a full covariance matrix $\Sigma$. By doing so, we model $p(\Y|\X)$ more accurately by not setting any restrictions on the form of $\Sigma$. As the diagonal elements of $\Sigma$ are considered individual uncertainty~\cite{Gal2016Uncertainty,NIPS2017_2650d608,pmlr-v48-gal16}, we further let \textit{off-diagonal} elements describe collaborative uncertainty. Diagonal element $\Sigma_{k,t,i,i}$ models individual uncertainty in the $k$-th prediction modal of the $i$-th agent at timestamp $t$. Off-diagonal element $\Sigma_{k,t,i,j}$ models collaborative uncertainty in the $k$-th prediction modal between the $i$-th and $j$-th agents at timestamp $t$. In this way, we obtain both individual and collaborative uncertainty by estimating the $\Sigma$ of $p(\Y|\X)$. Accordingly, we propose a probabilistic CU-aware regression framework with the following steps (see the visualization in Fig.~\ref{fig:three_step}).

\textbf{Step 1:} \emph{Choose a probability density function for the predictive distribution.} The chosen probability density function is formulated as $p(\Y|\X; \mu, \Phi, \Sigma)$, which includes a mean $\mu\in\R^{K\times 2T_{+}\times m }$ used to approximate the future trajectories, a covariance $\Sigma\in\R^{K\times T_{+} \times m\times m}$ used to quantify individual uncertainty and collaborative uncertainty, and some auxiliary parameters $\Phi$ used to describe the predictive distribution. Furthermore, we set covariance matrix $\Sigma_{k,t}$ as a full matrix instead of an identity or diagonal matrix.

\textbf{Step 2:} \emph{Design a regression model.} The regression model $\FF_{\w}(\cdot)$ is formulated as $\FF_{\w}(\cdot) = [\mu_{\w}(\cdot), \Phi_{\w}(\cdot),\Sigma_{\w}(\cdot)]$, where $\mu_{\w}(\cdot)$ is a trajectory predictor (T-Predictor) used to approximate the value of mean $\mu$, $\Sigma_{\w}(\cdot)$ is an uncertainty estimator (U-Estimator) estimating the value of covariance $\Sigma$, and $\Phi_{\w}(\cdot)$ is an auxiliary parameter estimator (AP-Estimator) estimating the values of some auxiliary parameters $\Phi$ when there is a need. Note that $\w$ is only used to indicate that the parameters of these three neural networks are trainable, not that they share the same set of parameters.

\textbf{Step 3:} \emph{Derive a loss function.} The loss function is derived from the $p(\Y|\X;\w)$ via the maximum likelihood estimation (MLE):
$\LL(\w)=-\log p(\Y|\X;\w)$. By minimizing this loss function, we update trainable parameters in $\FF_{\w}(\cdot) = [\mu_{\w}(\cdot), \Phi_{\w}(\cdot), \Sigma_{\w}(\cdot)]$.

In order to align the permutation of input observed trajectories and predicted trajectories, we implement $\mu_{\w}(\cdot)$ and $\Phi_{\w}(\cdot)$ by using two multi-layer perceptrons whose outputs are permutation-equivariant with the input data. Furthermore, since we model the individual and collaborative uncertainty via a covariance matrix $\Sigma$, we design a permutation-equivariant uncertainty estimator for $\Sigma_{\w}(\cdot)$, which outputs a positive definite matrix permutation-equivariant with the input data, in next subsection. Note that in the rest of this paper, we use $\widehat{\mu}$, $\widehat{\Phi}$ and $\widehat{\Sigma}$ to represent outputs of $\mu_{\w}(\cdot)$, $\Phi_{\w}(\cdot)$ and $\Sigma_{\w}(\cdot)$ respectively.

\subsubsection{Permutation-Equivariant Uncertainty Estimator}\label{subsec:PECUG}
In this subsection, we first illustrate the model structure of the permutation-equivariant uncertainty estimator, $\Sigma_{\w}(\cdot)$. Then, we provide proofs for the permutation equivariance and the positive definiteness of $\Sigma_{\w}(\X)$. To emphasize the output of $\Sigma_{\w}(\cdot)$ varies as the input $\X$ varies, here, we use $\Sigma_{\w}(\X)$ instead of $\widehat{\Sigma}$ to represent the output of $\Sigma_{\w}(\cdot)$.

\begin{figure}[t]
\begin{center}
\centerline{
\includegraphics[width=1.05\columnwidth]{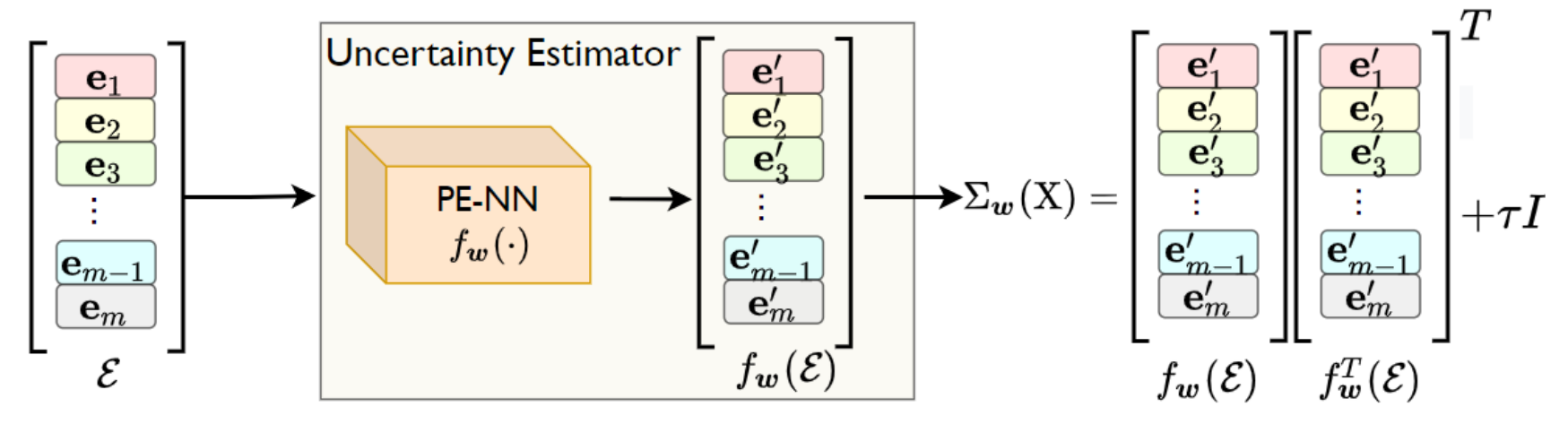}
 	}
\caption{\textbf{Permutation-equivariant uncertainty estimator.} This uncertainty estimator utilizes the input feature $\E$ to estimate the covariance $\Sigma_{\w}(\X)$. $\e_{i}$ is the feature generated on the basis of $\x_{i}$ and $f_{\w}(\cdot)$ is a permutation-equivariant neural network (PE-NN). $f_{\w}(\cdot)$ aims at projecting the feature $\E$ to $f_{\w}(\E)$ that is in a feature space where the individual uncertainty of $\x_i$ can be modeled by $\langle \e'_{i}$,$\e'_{i}\rangle$ and the collaborative uncertainty of $\x_i$ and $\x_j$ can be modeled by $\langle \e'_{i}$,$\e'_{j}\rangle$. $\tau$ is a positive real number and $I$ is an identity matrix.}
\label{fig:pecug}
\end{center}
\end{figure}

\textbf{Model structure.} The structure of $\Sigma_{\w}(\cdot)$ is illustrated in Fig.~\ref{fig:pecug}. Let $\E = [\e_1, \e_2, ..., \e_m]^{T}$ represent the input feature, which is permutation-equivariant with the input data $\X$, $\e_{i}$ represent the feature generated based on $\x_{i}$, and $f_{\w}(\cdot)$ be a permutation-equivariant neural network (PE-NN)\footnote{In practice, we implement it by a multi-layer perceptron.}. The intuition is that we employ the learning ability of $f_{\w}(\cdot)$ to generate feature $f_{\w}(\E) = [\e'_1,\e'_2, ..., \e'_m]^{T}$ based on $\E$. $f_{\w}(\E)$ is in a feature space where $\langle \e'_{i}$,$\e'_{i}\rangle$ models the individual uncertainty of $\y_i$, and where $\langle \e'_{i}$,$\e'_{j}\rangle$ models the collaborative uncertainty of $\y_i$ and $\y_j$. To make $\Sigma_{\w}(\X)$ a positive definite matrix, we generate it by adding a weighted identity matrix to the product of $f_{\w}(\E)$ and $f_{\w}^{T}(\E)$, which is formulated as: \begin{equation}
\setlength{\abovedisplayskip}{4pt}
\setlength{\belowdisplayskip}{4pt}
\Sigma_{\w}(\X) = f_{\w}(\E)f_{\w}^{T}(\E) + \tau I,
\label{eq:pecug}
\end{equation}
where $\tau$ is a positive real number and $I$ is an identity matrix.

\textbf{Theoretical properties of  $\Sigma_{\w}(\cdot)$.} The following two theorems show the uncertainty
estimator $\Sigma_{\w}(\cdot)$ promotes permutation-equivariance and positive definite properties.
\begin{theorem}
Given any input data $\X$, $\Sigma_{\w}(\X)$ is permutation-equivariant; that is,
\begin{equation*}
\setlength{\abovedisplayskip}{4pt}
\setlength{\belowdisplayskip}{4pt}
\Sigma_{\w}(P\X) = P\Sigma_{\w}(\X)P^{T},
\end{equation*}
where $P$ is a permutation matrix.
\end{theorem}

\begin{prf}
Because the input feature $\E$ is permutation-equivariant with the input data $\X$, we have:
\begin{equation}
\setlength{\abovedisplayskip}{4pt}
\setlength{\belowdisplayskip}{4pt}
\Sigma_{\w}(P\X) = f_{\w}(P\E)f_{\w}^{T}(P\E) + \tau I.
\label{eq:pfpe}
\end{equation}
Moreover, $f_{\w}(\cdot)$ is a permutation-equivariant neural network, $\tau$ is a positive real number and $I$ is an identity matrix, which means $f_{\w}(P\X) = Pf_{\w}(\X)$, $f^{T}_{\w}(P\X) = (Pf_{\w}(\X))^{T}$ and $P\tau IP^{T} = \tau I$. Therefore, (\ref{eq:pfpe}) can be reformulated as:
\begin{equation}
\setlength{\abovedisplayskip}{4pt}
\setlength{\belowdisplayskip}{4pt}
\Sigma_{\w}(P\X) = Pf_{\w}(\E)f^{T}_{\w}(\E)P^{T} + P\tau IP^{T}. 
\label{eq:pfpe2}
\end{equation}
On the basis of (\ref{eq:pecug}) and (\ref{eq:pfpe2}), we have:
\begin{equation*}
\setlength{\abovedisplayskip}{4pt}
\setlength{\belowdisplayskip}{4pt}
\Sigma_{\w}(P\X) = P\Sigma_{\w}(\X)P^{T}.
\end{equation*}
Therefore, $\Sigma_{\w}(\X)$ is permutation-equivariant with the input data $\X$.
\end{prf}

\begin{theorem}
Given any input data $\X$,  $\Sigma_{\w}(\X)$ is a positive definite matrix; that is,  $\bm{a}^{T}\Sigma_{\w}(\X)\bm{a} \geq 0$ for all $\bm{a} \in \R^{n}$; the equality holds if and only if $\bm{a} = \bm{0}$.
\end{theorem}
\begin{prf}
Based on (\ref{eq:pecug}), we have:
\begin{eqnarray*}
\bm{a}^{T}\Sigma_{\w}(\X)\bm{a} & = & \bm{a}^{T}f_{\w}(\E)f_{\w}^{T}(\E)\bm{a} + \bm{a}^{T}\tau I\bm{a} \\
& = & \bm{a}^{T}f_{\w}(\E)(\bm{a}^{T}f_{\w}(\E))^{T} + \tau\bm{a}^{T}\bm{a} \geq 0,
\end{eqnarray*}
where the equality holds if and only if $\bm{a} = \bm{0}$. As a result, $\Sigma_{\w}(\X)$ is a positive definite matrix.
\end{prf}

Compared with our previous work~\cite{tang2021collaborative}, this permutation-equivariant uncertainty estimator benefits uncertainty estimation in two ways: 1) making the generated covariance permutation-equivariant with the input data, which is more reasonable for real-world applications; 2) enabling the model to approximate the given distribution more accurately by generating more accurate covariance. See Subsection~\ref{subsec:syn_data} for related empirical results. 

\subsection{Special Case: Laplace CU-Aware Regression}\label{subsec:LCU}
In multi-agent trajectory forecasting, previous methods~\cite{liang2020learning,Gao_2020_CVPR,DBLP:conf/nips/KosarajuSM0RS19,tang2021collaborative} have found that the $\ell_1$-based loss function derived from the Laplace distribution usually leads to dominant prediction performances, because it is more robust to outliers. Therefore, here we show a special case of the framework proposed in Subsection~\ref{subsec:CUUEF} based on the multivariate Laplace distribution: \textit{Laplace CU-aware regression}.

\textbf{Probability density function.}
We follow the probabilistic formulation proposed in Subsection~\ref{subsec:CUUEF} and choose the probability density function as the multivariate Laplace distribution:
\begin{equation}
\setlength{\abovedisplayskip}{4pt}
\setlength{\belowdisplayskip}{4pt}
\begin{small}
p(\Y|\X) = \frac{2{\rm det} [\Gamma^{-1}]^{\frac{1}{2}} }{(2\pi)^{\frac{m}{2}}\lambda}\cdot\frac{K_{(\frac{m}{2}-1)}(\sqrt{\frac{2}{\lambda}(\Y-\mu)\Gamma^{-1}(\Y-\mu)^{T}})}{(\sqrt{\frac{\lambda}{2}(\Y-\mu)\Gamma^{-1}(\Y-\mu)^{T}})^{\frac{m}{2}}},
\label{eq:la_pdf}
\end{small}
\end{equation}
where $\Gamma$ is a positive definite matrix, $\Gamma^{-1}$ is the inverse matrix of $\Gamma$, ${\rm det} [\Gamma^{-1}]$ denotes the determinant of the matrix $\Gamma^{-1}$, $K_{(\frac{m}{2}-1)}(\cdot)$ denotes the modified Bessel function of the second kind with order $(\frac{m}{2}-1)$, and $\lambda$ is a positive real number. According to~\cite{1618702}, the covariance $\Sigma$ of the multivariable Laplace distribution is defined as:
\begin{equation}
\setlength{\abovedisplayskip}{4pt}
\setlength{\belowdisplayskip}{4pt}
\Sigma = \lambda\Gamma. 
\label{eq:co-of-lap}
\end{equation}
On the basis of (\ref{eq:co-of-lap}), for Laplace collaborative uncertainty, generating the covariance $\Sigma$ is equivalent to generating the $\Gamma$ matrix. Therefore, we employ the permutation-equivariant uncertainty estimator $\Sigma_{\w}(\cdot)$ mentioned in Subsection~\ref{subsec:CUUEF} to generate the $\Gamma$ matrix, and $\widehat{\Sigma}$ represents the generated $\Gamma$ matrix.

\begin{figure*}[t!]
\begin{center}
\centerline{\includegraphics[width = 2\columnwidth]{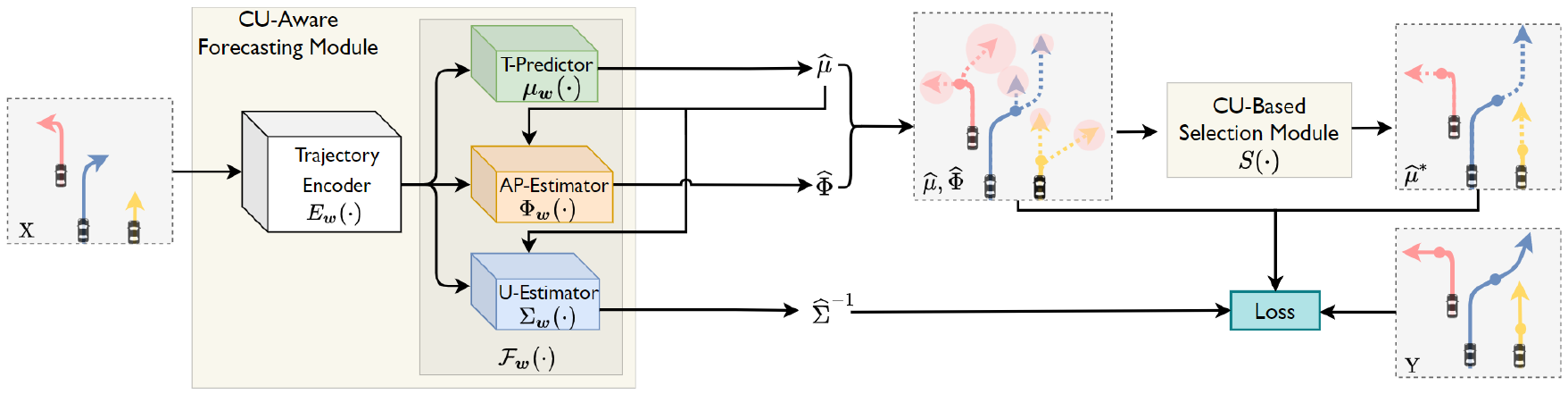}}
\caption{\textbf{CU-aware multi-agent multi-modal trajectory forecasting system.} This system consists of two modules: the CU-aware forecasting module and the CU-based selection module. The CU-aware forecasting module generating multi-modal predictions with the corresponding uncertainty. The CU-based selection module selects the optimal prediction generated by the forecasting module based on the estimated uncertainty. The system is trained in an end-to-end way with our proposed loss function (\ref{eq:lformft}).}
\label{fig:sframework}
\end{center}
\end{figure*}

\textbf{Model design.} Based on the discussion in Subsection~\ref{subsec:CUUEF}, we can approximate the value of mean $\mu$ via the T-Predictor $\mu_{\w}(\cdot)$, which is implemented by using a multi-layer perceptron. However, there are still two challenges for designing a model based on (\ref{eq:la_pdf}). First, if we use the neural network $\Sigma_{\w}(\cdot)$ to directly generate $\widehat{\Sigma} \in \R^{K \times T_{+} \times m \times m}$, each positive definite matrix $\widehat{\Sigma}_{k,t} \in \R^{m \times m}$ in $\widehat{\Sigma}$ needs to be inverted, which is computationally expensive and numerically unstable. Second, the modified Bessel function is intractable for neural networks to work with~\cite{1944A}.

For the first challenge, we make the permutation-equivariant uncertainty estimator $\Sigma_{\w}(\cdot)$ directly generate the $\widehat{\Sigma}^{-1}$. For the second challenge, inspired by~\cite{1618702}, we simplify (\ref{eq:la_pdf}) by utilizing the multivariate Gaussian distribution to approximate the multivariate Laplace distribution. We reformulate the multivariate Laplace distribution by introducing auxiliary variables. Let $\pPhi$ be a random variable from an exponential distribution\footnote{Note that the parameter $\lambda$ in (\ref{eq:la_pdf}), (\ref{eq:co-of-lap}) and (\ref{eq:expd}) are the same.}: 
\begin{equation}
\setlength{\abovedisplayskip}{4pt}
\setlength{\belowdisplayskip}{4pt}
    p(\pPhi|\X;\w) = \frac{1}{\lambda}e^{-\frac{\pPhi}{\lambda}}.
\label{eq:expd}
\end{equation}
Then (\ref{eq:la_pdf}) can be simplified as: 
\begin{equation*}
\setlength{\abovedisplayskip}{4pt}
\setlength{\belowdisplayskip}{4pt}
    p(\Y|\pPhi,\X;\w)=\frac{{\rm det}[\widehat{\Sigma}^{-1}]^{\frac{1}{2}}}{(2\pi\pPhi)^{\frac{m}{2}}}e^{-\frac{q(\Y)}{2\pPhi}},
\end{equation*}
where $q(\Y)=(\Y-\widehat{\mu})\widehat{\Sigma}^{-1}(\Y-\widehat{\mu})^{T}$. Furthermore, in this work, instead of drawing a value for $\pPhi$ from the exponential distribution, we use the AP-Estimator $\Phi_{\w}(\cdot)$ implemented by using a multi-layer perceptron to directly output a value $\pPhi^{*}$ for $\pPhi$. The intuition is that, in the training process of the regression model, the value of $p(\Y|\X;\w)$ is the conditional expectation of $\pPhi$ given $\X$ and $\Y$, which makes $p(\Y|\pPhi,\X;w)$ a function of $\pPhi$ whose domain is $\R^{+}$. Thus, there should exist an appropriate $\pPhi^{*}$ to make: 
\begin{equation*}
\setlength{\abovedisplayskip}{4pt}
\setlength{\belowdisplayskip}{4pt}
    p(\Y|\X;\w)=p(\Y|\pPhi^{*},\X;\w).
\end{equation*}
See the proof for the existence of $\pPhi^{*}$ in Appendix~\ref{ap:PLMD}. 

After getting the value of $\pPhi^{*}$ via $\Phi_{\w}(\cdot)$, we can get the parameterized form of $p(\Y|\X;\w)$ as: 
\begin{equation}
\setlength{\abovedisplayskip}{4pt}
\setlength{\belowdisplayskip}{4pt}
p(\Y|\X;\w) = \frac{{\rm det} [\widehat{\Sigma}^{-1}]^{\frac{1}{2}}}{(2\pi \widehat{\Phi})^{\frac{m}{2}}} e^{-\frac{q(\Y)}{2\widehat{\Phi}}}. 
\label{eq:sla_pdf}
\end{equation}
Moreover, the value of $\widehat{\Phi}$ can be used to approximate the value of $\lambda$, because the expected value of the exponentially distributed random variable $\pPhi$ is equal to $\lambda$. Therefore, the $\Sigma$ of chosen multivariate Laplace distribution can be approximated as
\begin{equation}
\setlength{\abovedisplayskip}{4pt}
\setlength{\belowdisplayskip}{4pt}
    \Sigma \approx \widehat{\Phi}\widehat{\Sigma}.
    \label{eq:cuap}
\end{equation}

Finally, we can get the regression model $\FF_{\w}(\cdot) = [\mu_{\w}(\cdot), \Phi_{\w}(\cdot), \Sigma_{\w}(\cdot)]$. Once $\widehat{\Sigma}$ and $\widehat{\Phi}$ are fixed, we can use them to compute $\Sigma$ as (\ref{eq:cuap}) to get individual uncertainty and collaborative uncertainty.

\textbf{Loss function.}
The log-likelihood function of (\ref{eq:sla_pdf}) is:
\begin{equation}
\setlength{\abovedisplayskip}{4pt}
\setlength{\belowdisplayskip}{4pt}
    \LL(\w) = \frac{1}{2} [\frac{q(\Y)}{\widehat{\Phi}}+m\log \widehat{\Phi} - \log {\rm det} [\widehat{\Sigma}^{-1}]].
    \label{eq:lcu1}
\end{equation}
Since $\widehat{\Sigma}^{-1}$ is a positive definite matrix, according to Hadamard's inequality~\cite{2004Inequalities}, we have:
\begin{equation*}
\setlength{\abovedisplayskip}{4pt}
\setlength{\belowdisplayskip}{4pt}
    \log {\rm det} [\widehat{\Sigma}^{-1}] \leq \sum\limits_{i=1}^{m}\log d_{ii},
\end{equation*}
where $d_{ii}$ is the diagonal element of $\widehat{\Sigma}^{-1}$.

Therefore, we have a lower bound of Equation~(\ref{eq:lcu1}) as:
\begin{equation}
\setlength{\abovedisplayskip}{4pt}
\setlength{\belowdisplayskip}{4pt}
    \LL_{\rm Lap\text{-}cu}(\w) = \frac{1}{2} [\frac{q(\Y)}{\widehat{\Phi}}+m\log \widehat{\Phi} - \sum\limits_{i=1}^{m}\log d_{ii}].
    \label{eq:lblcu1}
\end{equation}
The regression model $\FF_{\w}(\cdot) = [\mu_{\w}(\cdot), \Phi_{\w}(\cdot), \Sigma_{\w}(\cdot)]$ can be trained via minimizing this lower bound~(\ref{eq:lblcu1}).

In the next section, we will apply our proposed framework to the multi-modal multi-agent forecasting system based on this Laplace CU-aware regression model.

\section{CU-Aware Multi-Agent Multi-Modal Trajectory Forecasting System}
\label{sec:Example}
Multi-modal forecasting models are becoming increasingly important to the multi-agent trajectory forecasting task. They allow the system to predict multiple possible future trajectories for each agent, and thus better handle the stochasticity inherent in this task~\cite{liang2020learning, DBLP:journals/corr/abs-2008-08294,salzmann2020trajectron++}. However, existing multi-agent multi-modal trajectory forecasting systems are facing two challenges: 1) how to measure the uncertainty brought by the interaction module; and 2) how to utilize the uncertainty over each prediction to select the optimal prediction. In this section, in order to solve above two challenges, we apply our proposed framework to the system based on the Laplace CU-aware regression model, which leads to the CU-aware forecasting system (see Fig.~\ref{fig:sframework} for an illustration). In the following, we show the key component designs and the training strategy of this CU-aware forecasting system.

\subsection{Key Components}
A multi-agent multi-modal trajectory forecasting system usually consists of a forecasting module that generates multi-modal predictions and a selection module that selects the optimal prediction. Accordingly, for applying our proposed CU-aware regression framework to the multi-agent multi-modal trajectory forecasting system, we design a CU-aware forecasting module and a CU-based selection module.

\subsubsection{CU-Aware Forecasting Module}
\label{subsec:CU-FM}
To the best of our knowledge, most SOTA systems are equipped with a forecasting module based on the encoder-decoder architecture~\cite{MemoNet_2022_CVPR,xu2022GroupNet,bae2022npsn,9779572,gu2022stochastic,liang2020learning,Gao_2020_CVPR,Ye_2021_CVPR}. Therefore, we also design our CU-aware forecasting module based on such an architecture (see Fig.~\ref{fig:sframework} for the module illustration). In order to capture the interactions among agents in the input data, while aligning the permutation of input data and the generated feature, the trajectory encoder $E_{\w}(\cdot)$ is designed as a permutation-equivariant neural network with an interaction module. To enable the forecasting system to estimate the uncertainty corresponding to the predicted trajectories, we employ the  CU-aware regression model $\FF_{\w}(\cdot)$ shown in Subsection~\ref{subsec:LCU} as the decoder. The pipeline of the module is formulated as:
\begin{subequations}
\setlength{\abovedisplayskip}{4pt}
\setlength{\belowdisplayskip}{4pt}
\begin{align}
 \E = E_{\w}(\X),\nonumber 
\\
 \widehat{\mu} = \mu_{\w}(\E),\nonumber
\\
\widehat{\Phi} = \Phi_{\w}(\E,\widehat{\mu}),\nonumber
\\
\widehat{\Sigma}^{-1} = \Sigma_{\w}(\E,\widehat{\mu}).\nonumber
\end{align}
\end{subequations}

The trajectory encoder $E_{\w}(\cdot)$ generates the latent feature $\E$ containing the information about the historical trajectories of each agent. On the basis of the generated feature $\E$, the regression model $\FF_{\w}(\cdot)$ forecasts the future trajectories and estimates the corresponding uncertainty following this process: 1) the T-Predictor $\mu_{\w}(\cdot)$ in the $\FF_{\w}(\cdot)$ utilizes $\E$ to generate the multi-modal prediction $\widehat{\mu}\in \R^{K \times 2T_{+}\times m}$; 2) the AP-Estimator $\Phi_{\w}(\cdot)$ and U-Estimator $\Sigma_{\w}(\cdot)$ within the $\FF_{\w}(\cdot)$ generate $\widehat{\Phi}\in \R^{K\times T_{+} \times m}$ and $\widehat{\Sigma}^{-1}\in \R^{K\times T_{+} \times m\times m}$ to estimate the uncertainty of predictions in each prediction modal based on both $\E$ and $\widehat{\mu}$.

This CU-aware forecasting module enables the system to reflect the complete landscape of
uncertainty in multi-agent multi-modal trajectory forecasting, which leads to a more accurate approximation for the predictive distribution and makes the system more robust.

\subsubsection{CU-Based Selection Module}
For a system that predicts multiple future trajectories for each agent, it is natural to have a selection module that ranks those predictions according to their confidence levels. In the prior works~\cite{liang2020learning,zeng2021lanercnn,ye2021tpcn,Narayanan_2021_CVPR,Liu_2021_CVPR,Kim_2021_CVPR}, the selection module is trained to generate the highest confidence score for the prediction that is closest to the ground truth. However, there are two main drawbacks of this kind of selection module: 1) it is uninterpretable, since the ground truth data is usually with much randomness; 2) it ignores the uncertainty corresponding to each prediction when ranking the predicted trajectories, which may compromise the robustness of the model.

To overcome the above two flaws, we propose a novel CU-based selection module $S(\cdot)$, which does not require training. The proposed selection module will assign a highest ranking to the prediction with the lowest uncertainty as estimated by the CU-based framework, which can be formulated as:
\begin{equation*}
\setlength{\abovedisplayskip}{4pt}
\setlength{\belowdisplayskip}{4pt}
    \widehat{\mu}^{*} = S(\widehat{\mu},\widehat{\Phi}),
\end{equation*}
where $\widehat{\mu}$ is the prediction and $\widehat{\Phi}$ is the corresponding auxiliary parameter. Here, in the $k$-th modal, the uncertainty of $\widehat{\mu}_{k}$ is approximated by the auxiliary parameter $\widehat{\Phi}_{k}$ whose value is positively correlated with the uncertainty of $\widehat{\mu}_{k}$ according to (\ref{eq:cuap}). The use of $\widehat{\Phi}_{k}$ allows us to avoid computing the inverse of $\widehat{\Sigma}^{-1}_{k}$ matrix, which is computationally expensive and numerically unstable. To sum up, the CU-based trajectory selection module $S(\cdot)$ selects the $\widehat{\mu}_{k}$ corresponding to the lowest $\widehat{\Phi}_{k}$ as the $\widehat{\mu}^{*}$ for each agent.

Using our CU-based selection module leads to four gains: 1) enhancing the interpretability of the selection module in the forecasting system, because the uncertainty has a probabilistic interpretation: the variance of prediction; 2) enabling the selection module to utilize the uncertainty of each predicted trajectory; 3) avoiding the redundancy of training an extra neural network to generate the confidence score; 4) in Subsection~\ref{subsec:ab}, we empirically show that the CU-based selection module improves the performance of our forecasting system.

Compared with the system seen in our previous work~\cite{tang2021collaborative}, this CU-aware multi-agent multi-modal forecasting system is different in two perspectives: 1) the forecasting module is able to forecast future trajectories and estimate the corresponding uncertainty for multiple prediction modals instead of only one prediction modal; 2) the existence of CU-based selection module allows this system to select the optimal predicted trajectory based on the estimated uncertainty.

\subsection{Training Strategy}
The loss function used to train the CU-aware multi-agent multi-modal trajectory forecasting system is formulated as:
\begin{equation}
\setlength{\abovedisplayskip}{4pt}
\setlength{\belowdisplayskip}{4pt}
\LL_{\rm total}(\w) = \LL_{\rm Lap\text{-}cu}(\w) + \alpha \LL_{\rm AUTL}(\w),
\label{eq:lformft}
\end{equation}
where $\alpha$ is a hyperparameter, $\LL_{\rm Lap\text{-}cu}(\w)$ is (\ref{eq:lblcu1})\footnote{Note that, for making the predicted trajectories diverse, in (\ref{eq:lformft}) only $\widehat{\mu}^{*}$ and its corresponding $\widehat{\Phi}_{k}$ and $\widehat{\Sigma}_{k}$ are used to compute $\LL_{\rm Lap\text{-}cu}(\w)$.}, and $\LL_{\rm AUTL}(\w)$ is the auxiliary uncertainty training loss (AUTL), which is formulated as:
\begin{equation*}
\setlength{\abovedisplayskip}{4pt}
\setlength{\belowdisplayskip}{4pt}
    \LL_{\rm AUTL}(\w) = \sum\limits_{k=1}^{K}||\widehat{\Phi}_{k} - DE_{k}||_1,
\end{equation*}
where $DE_{k} = ||\widehat{\mu}_{k} - \Y||_2$ is the displacement error of the prediction in the $k$-th modal, $\widehat{\Phi}_{k}$ is the auxiliary parameter used to quantify the uncertainty of prediction in the $k$-th modal, and $\Y$ is the ground truth future trajectory. 

The intuition behind the AUTL is as follows. The values of the estimated uncertainty and the displacement error (DE) are positively correlated. The greater the corresponding uncertainty and DE are, the less reliable the predicted trajectory will be. Therefore, we are able to train our uncertainty estimator by minimizing the AUTL. Subsection~\ref{subsec:real_data} shows that the AUTL benefits the multi-modal forecasting system with our CU-aware regression framework, especially in terms of selecting the optimal predicted trajectory.

We can train the system in an end-to-end way with (\ref{eq:lformft}), as all of its components are differentiable.

\begin{figure*}[t!]
\begin{center}
\centerline{\includegraphics[width = 2.\columnwidth]{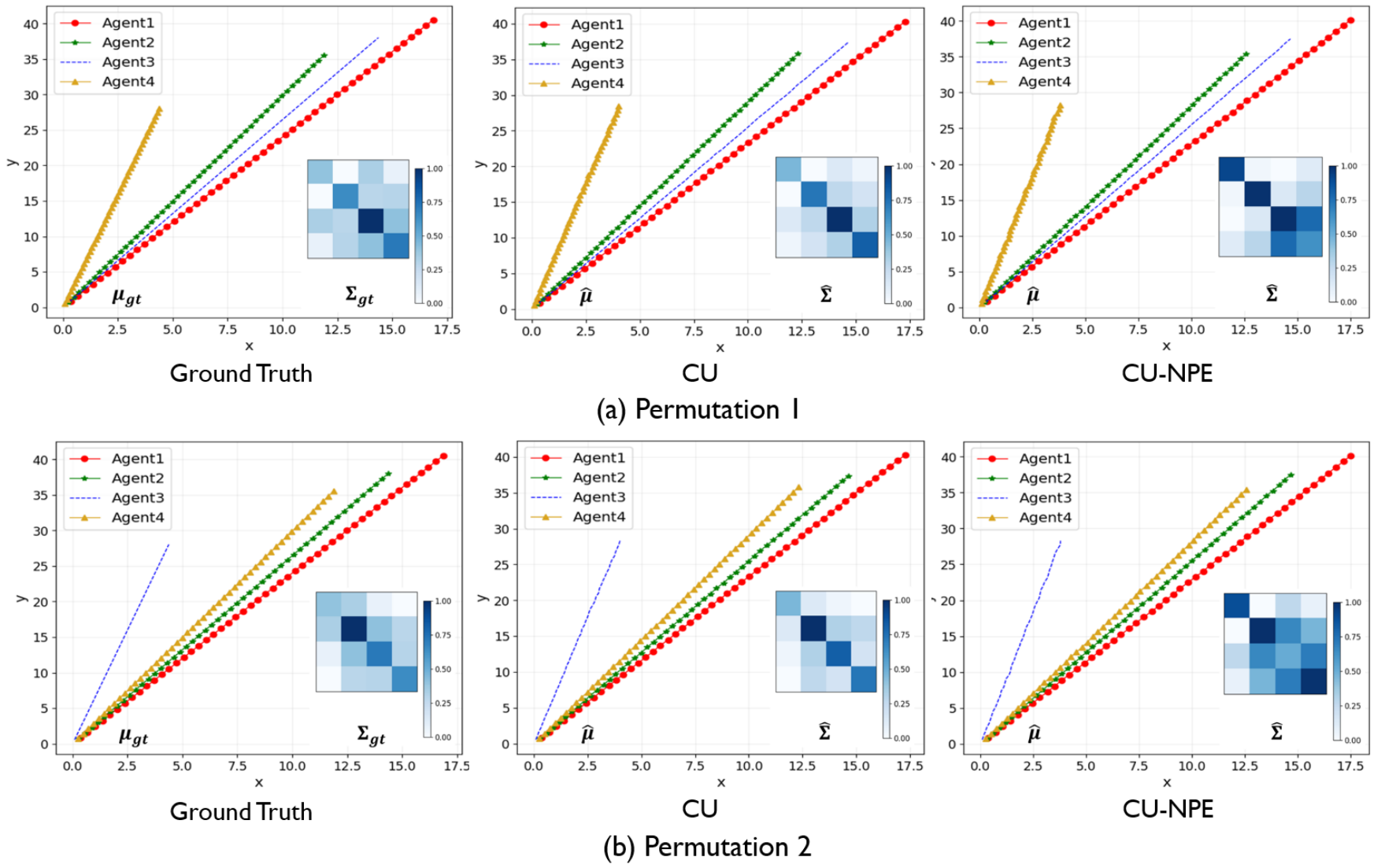}}
\caption{\textbf{Visual comparisons between outputs of the models and ground truth.} (a) and (b) show visualizations for the ground truth and the outputs of the models under two different permutations of the same set of inputs. Outputs of the CU model are consistently more accurate than outputs of the CU-NPE (None Permutation Equivariant) model. Further, outputs of the CU model are permutation-equivariant with the input data.}
\label{fig:permutation}
\end{center}
\end{figure*}

\section{Experiments and Analysis} \label{sec:Experiments}
We first use a self-generated synthetic dataset with a limited number of agents as the toy version of the real-world problem in Subsection~\ref{subsec:syn_data}. We use the simplified dataset to test our method's ability in capturing the distribution information of the input data that obeys the multivariate Laplace distribution, and to evaluate our proposed permutation-equivariant uncertainty estimator. We then conduct extensive experiments on two published benchmarks to prove the value of our proposed method in solving real-world problems in Subsection~\ref{subsec:real_data}.

\subsection{Toy Problem}
\label{subsec:syn_data}
We define a toy problem to validate the capability of the proposed framework in accurately estimating the probability distribution, and to evaluate our permutation-equivariant uncertainty estimator. The toy problem requires models to take the mutually correlated trajectories sampled from the given multivariate Laplace distribution as the input, and output the mean and covariance of this distribution.

As far as we know, in real-world datasets, we only have the ground truth for the predicted trajectory, which is the mean of the distribution, but we do not have access to the ground truth of the uncertainty, which is the covariance matrix of the distribution. Therefore, we generate a synthetic dataset with the ground truth for both the mean and the covariance matrix of the given multivariate Laplace distributions.

\subsubsection{Experimental Setup}
\textbf{Datasets.} We generate a synthetic dataset for the quaternary Laplace distribution. This dataset contains training, validation and test sets, which have 36000, 7000 and 7000 instances respectively. Each instance includes the trajectories of four agents consisting of the two-dimensional point coordinates of the four agents at 50 different timestamps. In each instance, the trajectories of the four agents are sampled from a quaternary joint Laplace distribution. Generation details are provided in Appendix~\ref{ap:TP-GDSD}.

\textbf{Implementation details.} We use the encoder-decoder network architecture for the toy problem. The neural network outputs the predicted mean and covariance matrix of the given quaternary Laplace distribution. Although the ground truth covariance matrix is known in the synthetic dataset, it is not used in training. We train the neural network with the previous uncertainty estimation methods~\cite{NIPS2017_2650d608,tang2021collaborative} and our proposed method respectively. Please find more details in Appendix~\ref{subsec:syn_model}.

\textbf{Metrics.} We adopt four metrics for evaluation: 1) the $\ell_2$ distances between the estimated mean and the ground truth mean, 2) the $\ell_1$ distances between the estimated $\Sigma$ matrix and the ground truth $\Sigma$ matrix, 3) the $\ell_1$ distances between the inverse of the estimated $\Sigma$ matrix and the inverse of the ground truth $\Sigma$ matrix, and 4) the KL divergence between the ground truth distribution and the estimated distribution. We provide the metrics computing details in Appendix~\ref{ap:TP-MCD}.

\begin{table}[t!]
\caption{Comparison with the prior uncertainty estimation framework on the synthetic dataset. \textit{CU} denotes the framework proposed in this paper, \textit{CU-NPE} denotes the framework proposed in~\cite{tang2021collaborative}, and \textit{IU Only} denotes the framework proposed in~\cite{NIPS2017_2650d608}. $\widehat{\mu}$: estimated mean. $\mu_{gt}$: ground truth mean. $\widehat{\Sigma}$: estimated covariance matrix. $\Sigma_{gt}$: ground truth covariance matrix.
$\ell_{2}$ of $\mu$: $||\widehat{\mu} - \mu_{gt}||_{2}$. $\ell_1$ of $\Sigma^{-1}$: $||\widehat{\Sigma}^{-1} - \Sigma^{-1}_{gt}||_{1}$. $\ell_1$ of $\Sigma$: $||\widehat{\Sigma} - \Sigma_{gt}||_{1}$. KL: KL divergence $D_{KL}(p_{g}(\X)||p_{e}(\X))$, where $p_{e}(\X)$ is the estimated distribution and $p_{g}(\X)$ is the ground truth distribution.}\label{tab:simulation_r}
\begin{center}
\begin{footnotesize}
\begin{small}{
\begin{tabular}{c|cccc}
\hline
& {$\ell_{2}$ of $\mu$} & {$\ell_1$ of $\Sigma^{-1}$} & {$\ell_1$ of $\Sigma$} & KL \\ \hline
IU Only~\cite{NIPS2017_2650d608} & 0.434  & 0.534    & 0.235    & 4.33  \\ 
CU-NPE\cite{tang2021collaborative}
& 0.376 & 0.502 & 0.209 & 3.15\\
CU
& \textbf{0.359}  & \textbf{0.271}  & \textbf{0.109}  & \textbf{2.11}  \\ 
\hline
\end{tabular}}
\end{small}
\end{footnotesize}
\end{center}
\end{table}

\subsubsection{Results}
\textbf{Quantitative results.} On the test set of our synthetic dataset, we compare the performance of our CU-aware regression framework with the permutation-equivariant uncertainty estimator (CU) and two previous uncertainty estimation frameworks: 1) the framework that only estimates individual uncertainty (IU Only)~\cite{NIPS2017_2650d608}; 2) the framework proposed in~\cite{tang2021collaborative}, which estimates uncertainty via the square-root-free Cholesky decomposition (CU-NPE). Table~\ref{tab:simulation_r} demonstrates that CU allows the model to more accurately estimate the mean and covariance matrix of the given distribution, which leads to a much lower KL divergence between the ground truth distribution and the estimated distribution. 

\textbf{Qualitative results.} Fig.~\ref{fig:permutation} illustrates the outputs of CU and CU-NPE under two different permutations of the same set of inputs, as well as their corresponding ground truths. These visualizations showcase: 1) CU consistently outputs more accurate covariance matrix than CU-NPE; 2) the covariance matrix generated by CU is permutation-equivariant with the input data.

\begin{figure*}[t!]
\begin{center}
\centerline{\includegraphics[width = 2.\columnwidth]{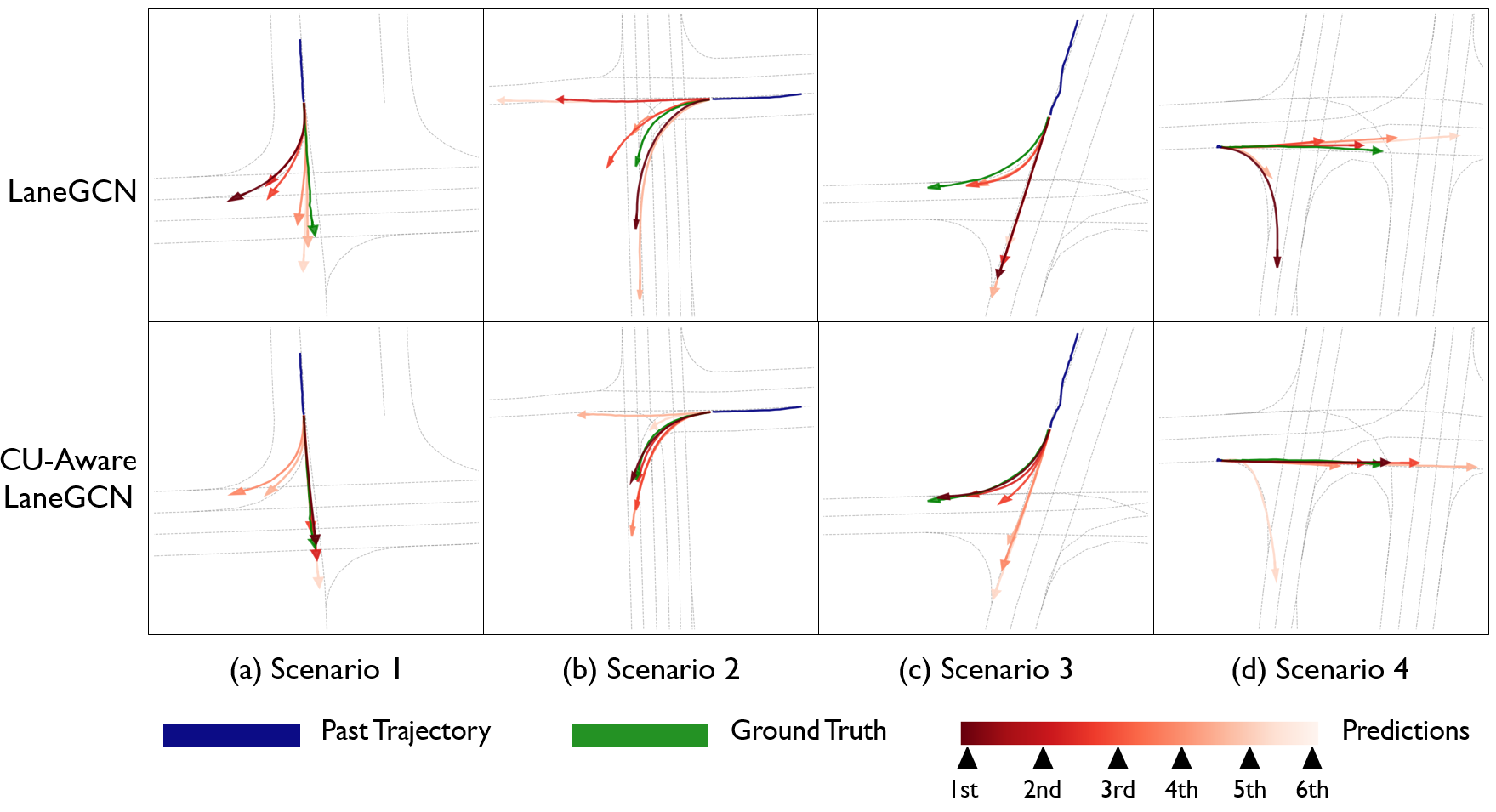}}
\caption{\textbf{Visualization of predicted trajectories.} (a), (b), (c) and (d) show the models' outputs in four different scenarios. The blue line is the past trajectory, the green line is the ground truth future trajectory, and the red lines are predictions generated by the models. The color bar indicates the reliability of each prediction: 1st stands for the most reliable prediction, 2nd for the second most reliable prediction, and so on. In all the four scenarios, predictions generated by the CU-aware LaneGCN is always closer to the ground truth than predictions generated by LaneGCN.}
\label{fig:cus}
\end{center}
\end{figure*}

\subsection{Real-world Problem}
\label{subsec:real_data}
\subsubsection{Experimental Setup}
\textbf{Datasets.} \emph{Argoverse}~\cite{Chang_2019_CVPR} and \emph{nuScenes}~\cite{nuscenes2019} are two widely used multi-agent trajectory forecasting benchmarks. \emph{Argoverse} has over 30000 scenarios collected in Pittsburgh and Miami. Each scenario is a sequence of frames sampled at 10 Hz. The sequences are split as training, validation and test sets, which have 205942, 39472 and 78143 sequences respectively. \emph{nuScenes} collects 1000 scenes in Boston and Singapore. Each scene is annotated at 2 Hz and is 20s-long. The prediction instances are split as training, validation and test sets, which have 32186, 8560 and 9041 instances respectively. For Argoverse, we forecast future trajectories for 3s based on the observed trajectories of 2s. For nuScenes, we forecast future trajectories for 6s based on the observed trajectories of 2s.

\begin{figure*}[t!]
\begin{center}
\centerline{\includegraphics[width = 2.\columnwidth]{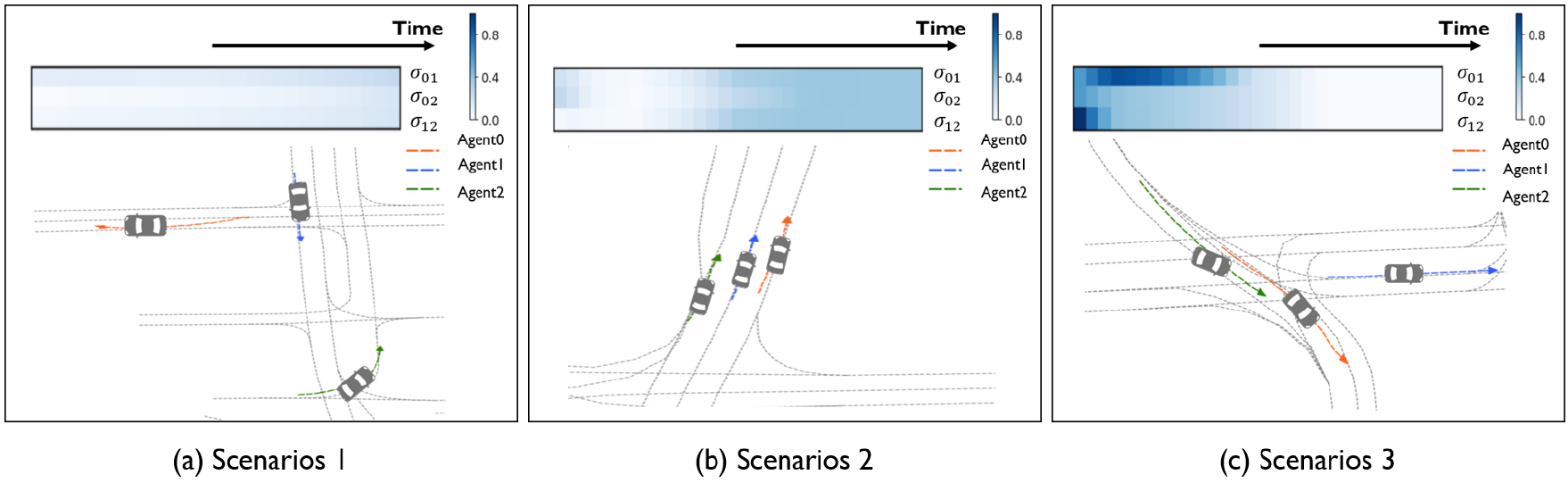}}
\caption{\textbf{Visualization of CU on the Argoverse dataset.} (a) In scenario 1, Agent 0, Agent 1 and Agent 2 are respectively located in completely different areas and heading towards different directions; $\sigma_{01}$, $\sigma_{02}$ and $\sigma_{12}$ are close to zero.
(b) In scenario 2, Agent 0, Agent 1 and Agent 2 are driving towards the same direction side by side. This type of scenario may generate complicated interactive information making $\sigma_{01}$, $\sigma_{02}$ and $\sigma_{12}$ show a non-monotonic change over time. (c) In scenario 3, Agent 0 first approaches the road on which Agent 1 is driving, and then moves away from that road, which causes $\sigma_{01}$ to first increase and then decrease. Agent 2 is far away from Agent 0 and Agent 1. Therefore, little new interactive information between Agent 1 and the other two agents would be generated, which makes $\sigma_{02}$ and $\sigma_{12}$ decrease over time.}
\label{fig:cu-v}
\end{center}
\end{figure*}

\textbf{Metrics.} We adopt seven widely used multi-agent trajectory forecasting metrics: $ADE$, $FDE$, $ADE_1$, $FDE_1$, $ADE_k$, $FDE_k$ and $Brier$-$FDE_k$. Firstly, $ADE$ is the Average Displacement Error, i.e., the average of point-wise $\ell_2$ distances between the prediction and the ground truth. Secondly, $FDE$ is the Final Displacement Error, i.e., the $\ell_2$ distance between the final points of the prediction and the ground truth. Thirdly, $ADE_1$/$FDE_1$ is $ADE$/$FDE$ of the selected optimal predicted trajectory Fourthly, $ADE_k$/$FDE_k$ is the minimum $ADE$/$FDE$ of predicted multi-modal trajectories. Finally, $Brier$-$FDE_k$ is the weighted $FDE_k$. (Metrics are in meters. $K=10$ on nuScenes and $K=6$ on Argoverse.)

\begin{table}[tb]
    \centering
    \caption{Comparison with SOTA methods on the Argoverse test set. Collaborative uncertainty estimation boosts performances.}
    \setlength{\tabcolsep}{1mm}{
    \begin{tabular}{c|ccccccc}
        \hline
        Methods & $ADE_1$ & $FDE_1$ & $ADE_k$ & $FDE_k$ & $Brier$-$FDE_k$\\
        \hline
        GOHOME~\cite{gilles2021gohome} & 1.69 & 3.65 & 0.94 & 1.45 & 1.98\\
        DenseTNT~\cite{gu2021densetnt}  & 1.68 & 3.63 & 0.88 & 1.28 & 1.98\\
        MMTransformer~\cite{liu2021multimodal}  & 1.77 & 4.00 & 0.84 & 1.34 & 2.03\\
        LaneRCNN~\cite{zeng2021lanercnn}  & 1.69 & 3.69 & 0.90 & 1.45 & 2.15\\
        LaneGCN~\cite{liang2020learning}  & 1.71 & 3.78 & 0.87 & 1.36 & 2.06\\
        \hline
        CU-aware LaneGCN & \textbf{1.62} & \textbf{3.55} & \textbf{0.83} & \textbf{1.26} & \textbf{1.95}\\
        \hline
    \end{tabular}}
    \label{tab:with_SOTA}
\end{table}

\textbf{Implementation details.} Here the model is implemented following the CU-aware multi-agent multi-modal trajectory forecasting system illustrated in Section~\ref{sec:Example}. We implement the encoder of LaneGCN~\cite{liang2020learning} and VectorNet~\cite{Gao_2020_CVPR}, which are two cutting-edge multi-agent trajectory forecasting models, as the trajectory encoder in our proposed model. We implement the encoder of LaneGCN based on their official released code.\footnote{LaneGCN official code: https://github.com/uber-research/LaneGCN} At the time of writing, there is no official release of the code for VectorNet yet, and hence we use our self-implemented VectorNet encoder in the experiments. We further use the multi-layer perceptrons to implement $\mu_{\w}(\cdot)$ and $\Phi_{\w}(\cdot)$. Finally, we implement $\Sigma_{\w}(\cdot)$ as the permutation-equivariant uncertainty estimator introduced in Subsection~\ref{subsec:CUUEF}.

\subsubsection{Results}
\textbf{Evaluation results on benchmark.} 
For evaluation, we implement our proposed framework based on LaneGCN~\cite{liang2020learning} as it is the SOTA model in multi-agent multi-modal trajectory forecasting. We name the modified LaneGCN that estimates collaborative uncertainty with our proposed CU-aware regression framework as CU-aware LaneGCN. We then compare the performance of CU-aware LaneGCN on the Argoverse trajectory forecasting benchmark with five SOTA methods of this benchmark: GOHOME~\cite{gilles2021gohome}, DenseTNT~\cite{gu2021densetnt}, MMTransformer~\cite{liu2021multimodal}, LaneRCNN~\cite{zeng2021lanercnn}, and LaneGCN~\cite{liang2020learning}. Table~\ref{tab:with_SOTA} demonstrates that the CU-aware LaneGCN especially outperforms all of the competing methods across all chosen metrics. Further, Fig.~\ref{fig:cus} \footnote{In order to keep figures concise, we only visualize the predictions and corresponding ground truths of one agent in each scenario.} visually compares the outputs of LaneGCN and CU-aware LaneGCN, which shows that CU-aware LaneGCN provides more accurate predictions and ranking results than LaneGCN. Therefore, estimating collaborative uncertainty enhances the SOTA prediction model.

\begin{table}[t]
\caption{Comparison with the previous collaborative uncertainty estimation framework in \textbf{\textit{single-modal trajectory forecasting}}. CU denotes the framework proposed in this paper and CU-NPE (None Permutation Equivariant) denotes the framework proposed in~\cite{tang2021collaborative}. $ADE$ and $FDE$ are metrics that evaluate model performance. On both Argoverse and nuScenes, the LaneGCN with CU always outperforms the LaneGCN with CU-NPE.}
\label{tab:single}
\begin{center}
\begin{tabular}{c|c|c|cc}
\hline
Method & Dataset & Framework &$ADE$ & $FDE$
\\
\hline
\multirow{2}{*}{} & \multirow{2}{*}{Argoverse}
&CU-NPE & 1.61 & 3.53\\
Single-Modal&&CU& \textbf{1.60} & \textbf{3.52} 
\\
\cline{2-5}
LaneGCN& \multirow{2}{*}{nuScenes}&CU-NPE & 5.27 & 10.55\\
&&CU& \textbf{5.21} & \textbf{10.52}
\\
\hline
\end{tabular}
\end{center}
\end{table}

\begin{table*}[t]
\caption{Ablation studies on the uncertainty estimation approach and the interaction module (\textbf{INT.}) in multi-agent multi-modal trajectory forecasting. \textbf{IU} denotes the approach only estimating individual uncertainty. \textbf{IU+CU} denotes our proposed approach that estimates both individual and collaborative uncertainty. On Argoverse and nuScenes, both LaneGCN and VectorNet with individual and collaborative uncertainty estimation surpass the ones with individual uncertainty estimation only. Collaborative uncertainty estimation has a larger impact on the performance of the model with an interaction module.}
\label{tab:ab_in_multi}
\begin{center}
\begin{tabular}{c|c|c|cc|c|cc|c|cc|c|cc|c}
\hline
\multirow{2}{*}{Method} &\multirow{2}{*}{Dataset} &\multirow{2}{*}{Int.} & \multicolumn{2}{c|}{$ADE_1$} & \multirow{2}{*}{$\Delta$} & \multicolumn{2}{c|}{$FDE_1$} & \multirow{2}{*}{$\Delta$}& \multicolumn{2}{c|}{$ADE_k$} & \multirow{2}{*}{$\Delta$} & \multicolumn{2}{c|}{$FDE_k$} & \multirow{2}{*}{$\Delta$}
\\
&  & &IU & IU+CU & & IU & IU+CU & & IU & IU+CU &  & IU & IU+CU\\
\hline
\multirow{4}{*}{\scriptsize LaneGCN}&\multirow{2}{*}{\scriptsize Argoverse}&  $\times$ & 1.81 & 1.75 & 3.3\% & 4.18 & 4.02 & 3.8\% & 0.91 & 0.90 & 1.1\% & 1.66 & 1.65 & 0.6\%\\
&&  $\surd$ & 1.33 & \textbf{1.28} & \textbf{3.8\%} & 2.92 & \textbf{2.78} & \textbf{4.8\%} & 0.70 & \textbf{0.69} & \textbf{1.4\%} & 1.03 & \textbf{1.01} & \textbf{1.9\%}
\\
\cline{2-15}
&\multirow{2}{*}{\scriptsize nuScenes}&$\times$ & 7.16 & 6.98 & 2.5\% & 14.74 & 14.34 & 2.7\% & 2.34 & 2.34 & 0.0\% & 3.95 & 3.91 & 1.0\%
\\
&& $\surd$ & 5.53 & \textbf{5.33} & \textbf{3.6\%} & 11.71 & \textbf{10.89} & \textbf{7.0\%} & 1.98 & \textbf{1.82} & \textbf{8.1\%} & 2.90 & \textbf{2.51} & \textbf{13.4\%}
\\
\hline
\hline
\multirow{4}{*}{\scriptsize VectorNet}&\multirow{2}{*}{\scriptsize Argoverse}& $\times$ & 1.80 & 1.80 & 0.0\% & 4.11 & 4.10 & 0.2\% & 1.52 & 1.51 & 0.7\% & 3.35 & 3.32 & 0.8\%
\\
&& $\surd$ & 1.53 & \textbf{1.50} & \textbf{1.9\%} & 3.37 & \textbf{3.31} & \textbf{1.8\%} & 1.32 & \textbf{1.27} & \textbf{4.0\%} & 2.77 & \textbf{2.66} & \textbf{4.0\%}
\\
\cline{2-15}
&\multirow{2}{*}{\scriptsize nuScenes}& $\times$ & 5.02 & 4.41 & 12.2\% & 12.07 & 10.76 & 10.9\% & 1.90 & 1.83 & 3.7\% & 3.84 & 3.67 & 4.4\%
\\
&&$\surd$ & 4.84 & \textbf{3.80} & \textbf{21.5\%} & 11.71 & \textbf{9.09} & \textbf{22.4\%} & 1.77 & \textbf{1.70} & \textbf{4.0\%} & 3.50 & \textbf{3.22} & \textbf{8.0\%}
\\
\hline
\end{tabular}
\end{center}
\end{table*}
\label{as_A2A}
\begin{table}[t]
\caption{Ablation studies on the CU-based selection module (CUSelect). For the model without the CU-based selection module, we train an extra neural network to generate the confidence score for it. On both Argoverse and nuScenes, CU-aware models (CU-aware LaneGCN/VectorNet) with the CUSelect always outperform models without it.}
\label{tab:ab_cus}
\setlength{\tabcolsep}{1.2mm}{
\begin{center}
\begin{tabular}{c|c|c|cccc}
\hline
Method & Dataset & CUSelect & $ADE_1$& $FDE_1$&$ADE_k$ & $FDE_k$
\\
\hline
&\multirow{2}{*}{Argoverse}&  $\times$ & 1.45 & 3.26 & 0.70 & 1.03\\
CU-aware&&  $\surd$ & \textbf{1.28} & \textbf{2.78} & \textbf{0.69} & \textbf{1.01}
\\
\cline{2-7}
LaneGCN&\multirow{2}{*}{nuScenes}&$\times$ & 6.22 & 12.64 & 1.83 & 2.61
\\
&& $\surd$ &  \textbf{5.33} & \textbf{10.89} & \textbf{1.82} & \textbf{2.51}
\\
\hline
\hline
&\multirow{2}{*}{Argoverse}& $\times$ &1.55 & 3.42 & 1.34 & 2.82
\\
 CU-aware&& $\surd$ & \textbf{1.50} & \textbf{3.31} & \textbf{1.26} & \textbf{2.67}
\\
\cline{2-7}
VectorNet&\multirow{2}{*}{ nuScenes}& $\times$ & 5.89 & 14.14 & 1.88 & 3.82
\\
&&$\surd$ & \textbf{3.80} & \textbf{9.09} & \textbf{1.70} & \textbf{3.22}
\\
\hline
\end{tabular}
\end{center}}
\end{table}

\textbf{Evaluation results for single-modal forecasting.} To compare our proposed CU-aware regression framework (CU) and the previous collaborative uncertainty estimation framework (CU-NPE) proposed in~\cite{tang2021collaborative} under the multi-agent trajectory forecasting setting. We generalize LaneGCN to the single-modal scenario by letting the forecasting module only predict one possible future trajectory for each agent and removing the selection module. Table~\ref{tab:single} shows that, on the test set of both Argoverse and nuScenes, the single-modal LaneGCN with CU always outperforms it with CU-NPE. We speculate that this improvement may come from the permutation-equivariant uncertainty estimator, which makes the uncertainty estimation more accurate.

\textbf{Visualization of collaborative uncertainty.} To visually understand which factor influences the value of CU in multi-agent multi-modal trajectory forecasting, Fig.~\ref{fig:cu-v} illustrates the visualization results generated by our model. We visualize three scenarios. Each scenario includes three agents' trajectories (i.e., orange, blue and green dash lines) and their corresponding CU values changing over the last 30 frames (i.e., the heatmap, where $\sigma_{ij}$ denotes the CU of agent $i$ and agent $j$). The visualization results show that the value of CU in multi-agent multi-modal trajectory forecasting is highly related to the interactive information among agents.\footnote{Note that to keep the visualization results concise, for each scenario, we only visualize the optimal trajectory and its corresponding CU.}

\subsubsection{Ablation Study}
\label{subsec:ab}
We study: 1) how different approaches of uncertainty estimation would impact the prediction model; 2) how the proposed CU-based selection module would affect the CU-aware multi-modal trajectory forecasting system; 3) how the proposed auxiliary uncertainty training loss (AUTL) would influence the performance of the CU-aware regression framework in multi-modal trajectory forecasting. In this part, we adopt LaneGCN/VectorNet as our trajectory encoder for proving that our proposed method can be used as a plug-in module to improve the performance of existing models for multi-agent multi-modal trajectory forecasting. Here, experiments are conducted on the validation sets of the Argoverse and nuScenes benchmarks.

\textbf{Effect of different uncertainty estimation approaches.} In the multi-agent multi-modal trajectory forecasting task, on the basis of LaneGCN/VectorNet, we consider two approaches of uncertainty estimation: 1) estimating individual uncertainty only (IU); 2) estimating both collaborative and individual uncertainty (IU + CU). Table~\ref{tab:ab_in_multi} demonstrates that estimating both individual and collaborative uncertainty (IU + CU) is consistently better than estimating individual uncertainty only (IU). These results reflect that our collaborative uncertainty estimation (i.e., the CU-aware regression framework) can function as a plugin module to improve the prediction performance of existing models.

\textbf{Effect of the CU-based selection module.} Here, we empirically study how the proposed CU-based selection module influences the CU-aware multi-agent multi-modal trajectory forecasting system. For the model without the CU-based selection module, we train an extra neural network to generate the confidence score for each predicted trajectory. Table~\ref{tab:ab_cus} shows that, the forecasting model with the CU-based selection module always provides more accurate results than the model without it, which indicates that our CU-based selection module improves the performance of the CU-aware forecasting model.

\begin{table}[t!]
\caption{Ablation studies on auxiliary uncertainty training loss (AULT) in CU-aware multi-modal trajectory forecasting models (CU-aware LaneGCN/VectorNet). $\Delta$ represents the improvement that AUTL brings to the forecasting model. On both Argoverse and nuScenes, AUTL benefits the prediction model, and it impacts $ADE_1/FDE_1$ significantly, more so than it impacts $ADE_K/FDE_K$.}
\label{tab:AUTL}
\begin{center}
\begin{tabular}{c|c|c|cc|c}
\hline
\multirow{2}{*}{Method} &\multirow{2}{*}{Dataset} &\multirow{2}{*}{Metrics} & \multicolumn{2}{c|}{AULT} & \multirow{2}{*}{$\Delta$}
\\
&  & &$\times$ & $\surd$ & \\
\hline
\multirow{3}{*}{}&\multirow{4}{*}{ Argoverse}&  $ADE_K$ & 0.70 & \textbf{0.69} & 2.6\%
\\
 &&  $ADE_1$ & 1.42 & \textbf{1.28} & \textbf{9.6\%}
\\
\cline{3-6}
&&  $FDE_K$ & 1.06 & \textbf{1.01} & 5.1\%
\\
CU-aware&&  $FDE_1$ & 3.14 & \textbf{2.78} & \textbf{11.4\%}
\\
\cline{2-6}
LaneGCN&\multirow{4}{*}{nuScenes}&  $ADE_K$ & 1.84
 & \textbf{1.82}
 & 1.4\%
\\
\multirow{3}{*}{}&&  $ADE_1$ & 11.40
 & \textbf{5.33}
 & \textbf{53.3\%}
\\
\cline{3-6}
&&  $FDE_K$ & 2.57
 & \textbf{2.51}
 & 2.2\%
\\
&&  $FDE_1$ & 23.39
 & \textbf{10.89}
 & \textbf{53.4\%}
\\
\hline
\hline
\multirow{3}{*}{}&\multirow{4}{*}{ Argoverse}&  $ADE_K$ & 1.30
 & \textbf{1.27}
 & 3.0\%
\\
&&  $ADE_1$ & 2.40
 & \textbf{1.50}
& \textbf{37.4\%}
\\
\cline{3-6}
&&  $FDE_K$ & 2.75
 & \textbf{2.66}
 & 3.3\%
\\
CU-aware&&  $FDE_1$ & 5.50
 & \textbf{3.31}
 & \textbf{39.8\%}
\\
\cline{2-6}
VectorNet&\multirow{4}{*}{nuScenes}&  $ADE_K$ & 1.79
 & \textbf{1.70}
 & 5.3\%
\\
\multirow{3}{*}{}&&  $ADE_1$ & 4.86
 & \textbf{3.80}
 & \textbf{21.8\%}
\\
\cline{3-6}
&&  $FDE_K$ & 3.55
 & \textbf{3.22}
 & 9.3\%
\\
&&  $FDE_1$ & 11.79
 & \textbf{9.09}
 & \textbf{22.9\%}
\\
\hline
\end{tabular}
\end{center}
\end{table}
\label{as_A2A}

\textbf{Effect of the auxiliary uncertainty training loss (AUTL).} The ablation study results on the proposed AUTL are shown in Table~\ref{tab:AUTL}. Results show that CU-aware models (CU-aware LaneGCN/VectorNet) trained with AUTL always outperform CU-aware models trained without it on all chosen metrics, and that AULT makes a larger improvement in $ADE_1/FDE_1$ than in $ADE_K/FDE_K$. These results show a proof that the proposed AUTL benefits CU-aware multi-agent multi-modal trajectory forecasting models, especially in terms of choosing the optimal predicted trajectory. 

\begin{figure*}[t]
\begin{center}
\centerline{
\includegraphics[width=1.8\columnwidth]{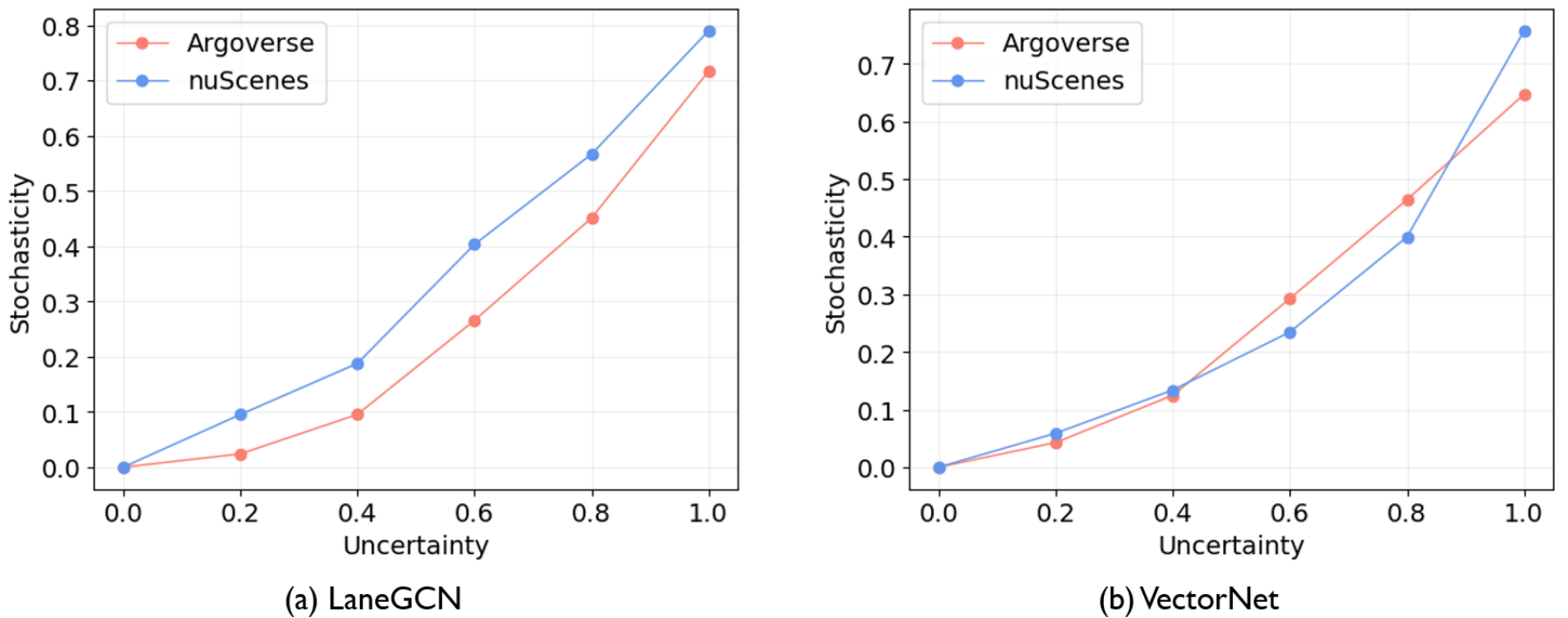}}
\caption{\textbf{Relationship between future stochasticity and the prediction uncertainty (i.e., the stochasticity-uncertainty relationship)} for (a) LaneGCN: LaneGCN on the Argoverse dataset and the nuScenes dataset, (b) VectorNet: VectorNet on the Argoverse dataset and the nuScenes dataset. These results show a positively correlated stochasticity-uncertainty relationship.}
\label{fig:diver-un}
\end{center}
\end{figure*}
\begin{figure*}[t!]
\begin{center}
\centerline{\includegraphics[width = 1.8\columnwidth]{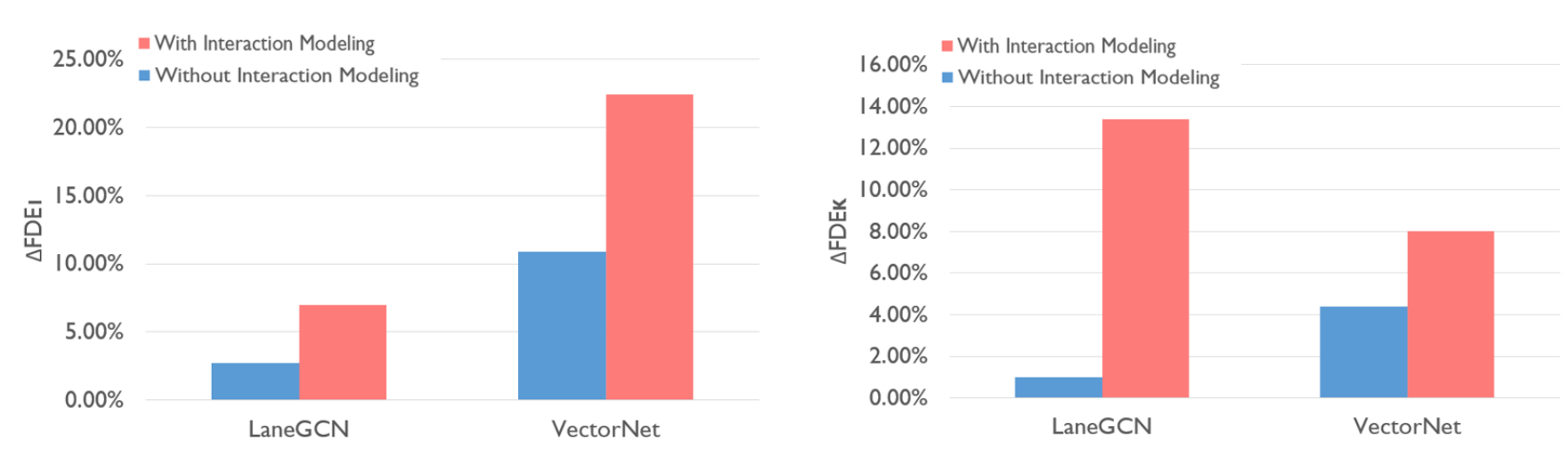}}
\caption{\textbf{CU estimation gains more when a prediction model includes interaction modeling.} Blue and red bars reflect the gains from CU estimation for a prediction model with/without an interaction module respectively. $\Delta FDE_1$/$\Delta FDE_k$ are the improvement rate of $FDE_1$/$FDE_k$ between versions with/without CU estimation respectively. Data of this figure comes from experiments on the nuScenes dataset shown in Table~\ref{tab:ab_in_multi}.}
\label{fig:comp_in}
\end{center}
\end{figure*}

\section{Discussion: What Causes the Uncertainty in Multi-Agent Multi-Modal Trajectory Forecasting}
\label{sec:discussion}
In Section~\ref{sec:Experiments}, our experimental results show that the CU-aware regression framework can be used to aid models in approximating the distribution information of given data and in improving prediction performance. In this section, we leverage the proposed CU-aware regression framework to study the cause of the uncertainty in multi-agent multi-modal trajectory forecasting, particularly the cause of our proposed collaborative uncertainty.

\subsection{Cause of Uncertainty}
\label{subsec:D_and_Un}
As mentioned in Section~\ref{sec:Related_Works} and Section~\ref{sec:Methodology}, potential future trajectories of multiple agents are inherently stochastic. We argue that future stochasticity contributes to the uncertainty in multi-agent multi-modal trajectory forecasting. 

To support this argument, we empirically study the relationship between future stochasticity and the prediction uncertainty (i.e., the stochasticity-uncertainty relationship) of multi-agent multi-modal trajectory forecasting models. In our experiments, the prediction uncertainty is quantified by our proposed Laplace CU-aware regression framework. For the scale of future stochasticity, according to \cite{DBLP:journals/corr/abs-1911-12736} and \cite{yuan2020dlow}, the prediction diversity of a forecasting model is positively correlated with the stochasticity of given agents' potential future trajectories. Therefore, we use the prediction diversity of multi-agent multi-modal trajectory forecasting systems to estimate the future stochasticity scale, which can be quantified by (\ref{eq:DS}).

\begin{equation}
\setlength{\abovedisplayskip}{4pt}
\setlength{\belowdisplayskip}{4pt}
    {\rm Stochasticity} =  m( \frac{\sum\limits_{i=1}^{K}(\widehat{\mu}_i - \overline{\widehat{\mu}})^{2}}{K-1}),
    \label{eq:DS}
\end{equation}
where $\widehat{\mu}=[\widehat{\mu}_{1}, \widehat{\mu}_{2}, ..., \widehat{\mu}_{K}]$ is a multi-modal prediction result, $K$ is the number of the prediction modal, $\widehat{\mu}_{k} \in \R^{2T_{+}}$, $\overline{\widehat{\mu}} \in \R^{2T_{+}}$ is the mean of elements in $\widehat{\mu}$, and $m(\cdot)$ symbolizes the computation of element-wise average of a vector.

We visually analyze the stochasticity-uncertainty relationship for LaneGCN as well as VectorNet on the Argoverse and nuScenes datasets. See visualization results in Fig.~\ref{fig:diver-un}. These results show a positive correlation between future stochasticity and prediction uncertainty in multi-agent multi-modal trajectory forecasting models. Therefore, the stochasticity of given agents' potential future trajectories contributes to the existence of uncertainty in multi-agent multi-modal trajectory forecasting.

\subsection{Cause of Collaborative Uncertainty}
\label{subsec:NCU}
In this subsection, we focus on the cause of our proposed collaborative uncertainty. As mentioned in Section~\ref{sec:problem_formulation}, we can divide multi-agent multi-modal trajectory forecasting models into two types: the individual model and the collaborative model. The individual model predicts the future trajectory and the corresponding uncertainty for each agent independently, while the collaborative model leverages an interaction module to explicitly capture the interactions among multiple agents, which makes all the predicted trajectories correlated. The correlations among predicted trajectories can bring extra uncertainty to the model; in other words, we consider that the interaction module in a forecasting model leads to collaborative uncertainty.

To validate this, we empirically compare the impact of collaborative uncertainty estimation on an individual model versus a collaborative model. In our experiments, we use two cutting-edge multi-agent trajectory forecasting models, LaneGCN~\cite{liang2020learning} and VectorNet~\cite{Gao_2020_CVPR}. Furthermore, the improvement rate of $ADE_1$, $FDE_1$, $ADE_k$, and $FDE_k$ between models with/without an interaction module are represented by $\Delta ADE_1$, $\Delta FDE_1$, $\Delta ADE_k$, and $\Delta FDE_k$ respectively. As Fig.~\ref{fig:comp_in} illustrates, when we remove the interaction module from LaneGCN/VectorNet, collaborative uncertainty estimation brings much less gain to LaneGCN/VectorNet (see Table~\ref{tab:ab_in_multi} for more results of this experiment). As a result, the interaction module causes the collaborative uncertainty in multi-agent multi-modal trajectory forecasting.
\section{Conclusions} \label{sec:Conclusions}

This work proposes a novel collaborative uncertainty (CU) aware regression framework for multi-agent multi-modal trajectory forecasting, which contains a novel permutation-equivariant uncertainty estimator. The framework's key novelty is twofold. Firstly, it conceptualizes and models the collaborative uncertainty introduced by interaction modules. Secondly, it enables the multi-agent multi-modal trajectory forecasting system to rank multi-modal predictions based on the uncertainty over each prediction. Results of extensive experiments demonstrate the ability of this CU-aware regression framework in boosting the performance of SOTA forecasting systems, especially collaborative-model-based systems. With further experiments, we reveal a positive correlation between future stochasticity and prediction uncertainty in multi-agent multi-modal trajectory forecasting systems. We believe that this work, as well as the framework it presents, will guide the development of more reliable and safer forecasting systems in the near future. 

\ifCLASSOPTIONcompsoc
  \section*{Acknowledgments}
\else
  \section*{Acknowledgment}
\fi
This research is partially supported by the National Key R\&D Program of China under Grant 2021ZD0112801, National Natural Science Foundation of China under Grant 62171276, the Science and Technology Commission of Shanghai Municipal under Grant 21511100900 and CCF-DiDi GAIA Research Collaboration Plan 202112.

\ifCLASSOPTIONcaptionsoff
  \newpage
\fi

\bibliographystyle{IEEEtran}
\bibliography{mybib}

\begin{IEEEbiography}[{\includegraphics[width=1in,height=1.25in,clip,keepaspectratio]{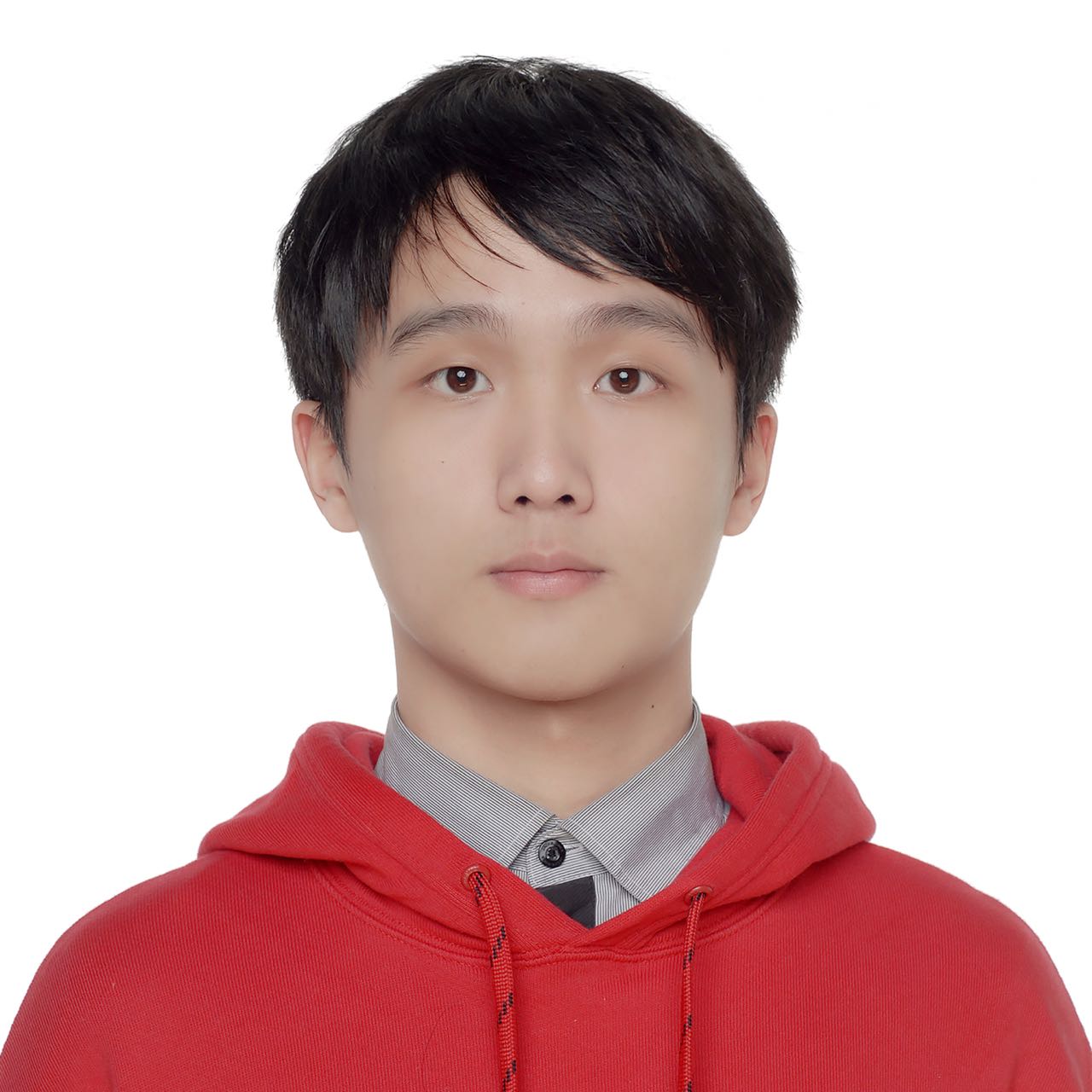}}]{Bohan Tang} recieved the B.E. degree in information engineering from Shanghai Jiao Tong University, Shanghai, China, in 2021. He is working toward the Ph.D. degree at the Oxford-Man Institute and the Department of Engineering Science in University of Oxford since 2021. His research interests include machine learning, graph neural network, graph learning, hypergraph learning and uncertainty estimation. 
\end{IEEEbiography}

\begin{IEEEbiography}[{\includegraphics[width=1in,height=1.25in,clip,keepaspectratio]{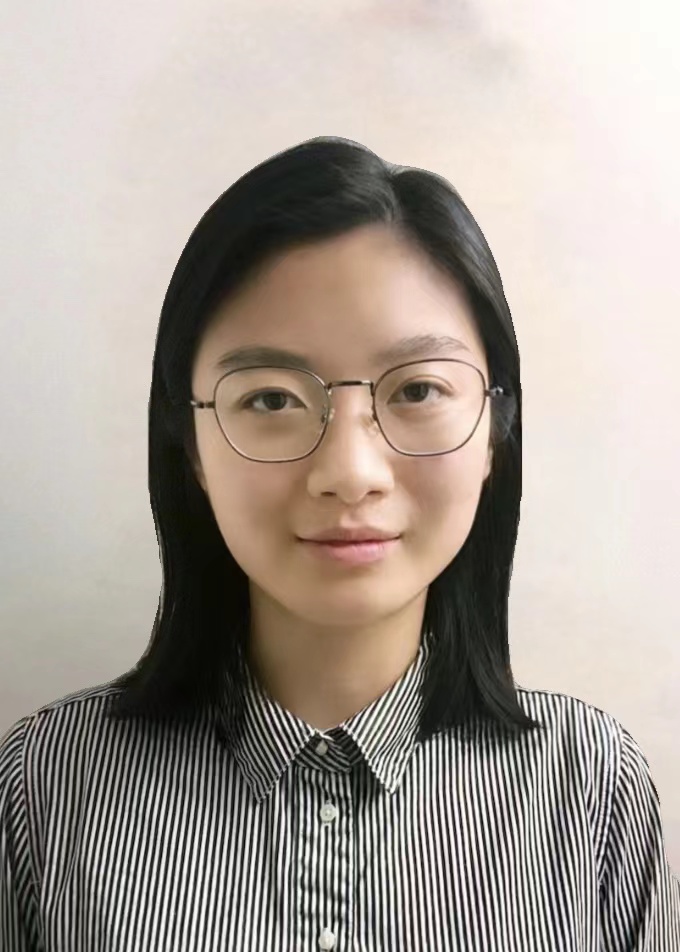}}]{Yiqi Zhong}, at the time of writing, is a Ph.D. candidate in Computer Science at the University of Southern California, where she received an M.S. in Computer Science in 2018. She received her bachelor's degree from Dalian Maritime University, China, in 2016. Affiliated to the USC Computer Graphics and Immersive Technologies (CGIT) laboratory (cgit.usc.edu), her research focuses on computer vision, multimodal learning, and multi-agent trajectory prediction. 
\end{IEEEbiography}

\vspace{-1mm}

\begin{IEEEbiography}[{\includegraphics[width=1in,height=1.25in,clip,keepaspectratio]{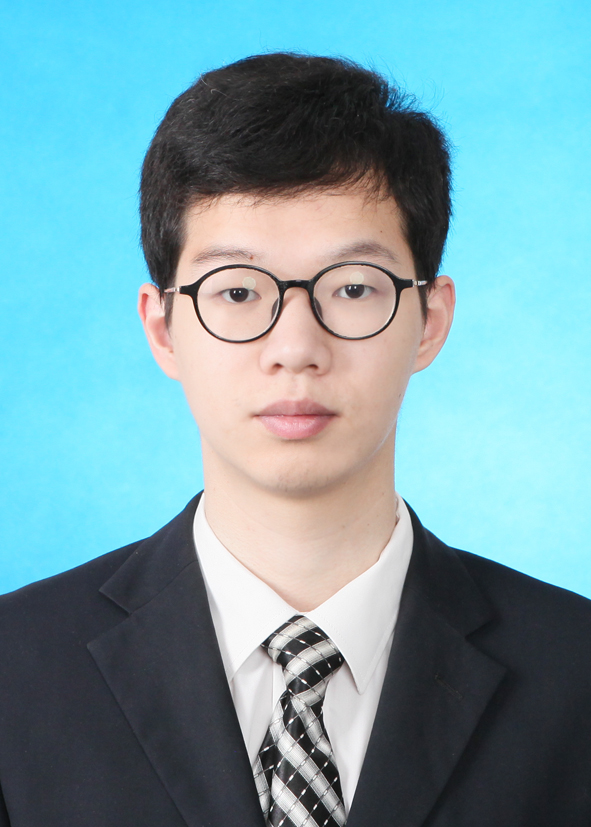}}]{Chenxin Xu} received a B.E. degree in Information Engineering from Shanghai Jiao Tong University, Shanghai, China, in 2019. He is working toward a joint Ph.D. degree at the Cooperative Medianet Innovation Center in Shanghai Jiao Tong University and at Electrical and Computer Engineering in National University of Singapore since 2019. His research interests include computer vision, machine learning, graph neural network, and multi-agent prediction. 
\end{IEEEbiography}
\vspace{-1mm}

\begin{IEEEbiography}[{\includegraphics[width=1in,height=1.7in,clip,keepaspectratio]{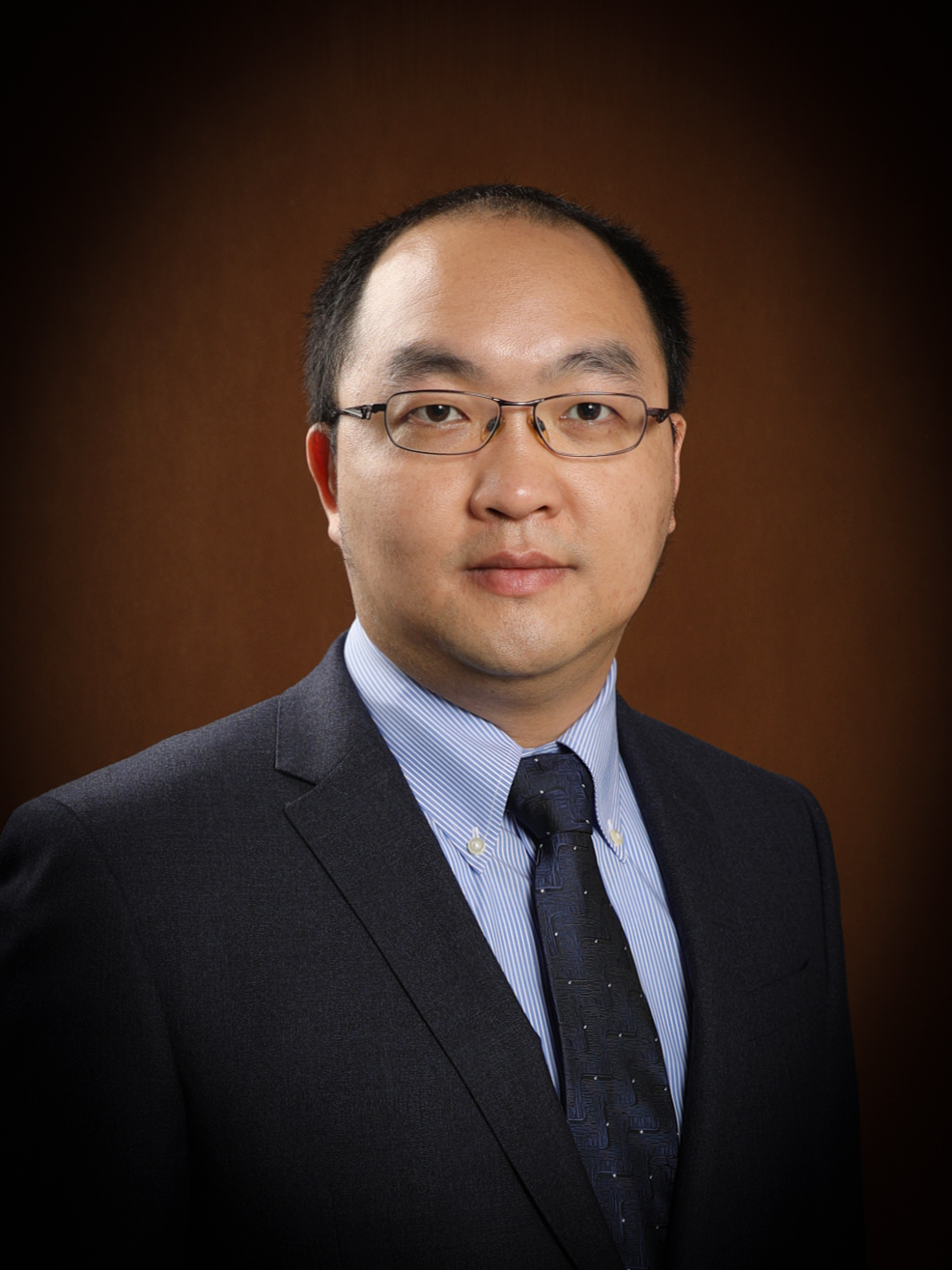}}]{Wei-Tao Wu} is a professor of School of Mechanical Engineering at Nanjing University of Science and Technology. Before joining Nanjing University of Science and Technology, he was a postdoc research associate at Department of Biomedical Engineering of Carnegie Mellon University. Dr. Wu received his B.S. degree from Xi’an Jiaotong University in 2011 and his Ph.D. degree from the Department of Mechanical Engineering of Carnegie Mellon University in 2015. Dr. Wu’s research interests lie in fluid mechanics, heat transfer and physics-informed machine learning. Within recent 5 years, Dr. Wei-Tao Wu has authored or co-authored over 50 peer-reviewed journal papers, including 3 journal cover/featured papers. Dr. Wei-Tao Wu has also served as a member of Physics and Aerodynamics Committee of Chinese Aerodynamics Research Society. 
\end{IEEEbiography}

\begin{IEEEbiography}[{\includegraphics[width=1in,height=1.7in,clip,keepaspectratio]{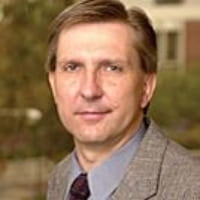}}]{Ulrich Neumann} is a professor in the Computer Science Department at the University of Southern California with a joint appointment in the Ming Hsieh Department of Electrical and Computer Engineering. He has an MSEE from the State University of New York at Buffalo and a Ph.D. in Computer Science from the University of North Carolina at Chapel Hill. His research interests lie in the intersection of computer vision and computer graphics.
\end{IEEEbiography}

\begin{IEEEbiography}[{\includegraphics[width=1in,height=1.25in,clip,keepaspectratio]{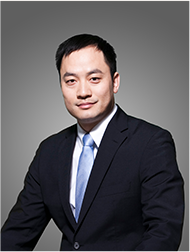}}]{Yanfeng Wang} received the B.E. degree in information engineering from the University of PLA, Beijing, China, and the M.S. and Ph.D. degrees in business management from the Antai College of Economics and Management, Shanghai Jiao Tong University, Shanghai, China. He is currently the Vice Director of the Cooperative Medianet Innovation Center and also the Vice Dean of the School of Electrical and Information Engineering, Shanghai Jiao Tong University. His research interests mainly include media big data and emerging commercial applications of information technology.
\end{IEEEbiography}

\begin{IEEEbiography}[{\includegraphics[width=1in,height=1.25in,clip,keepaspectratio]{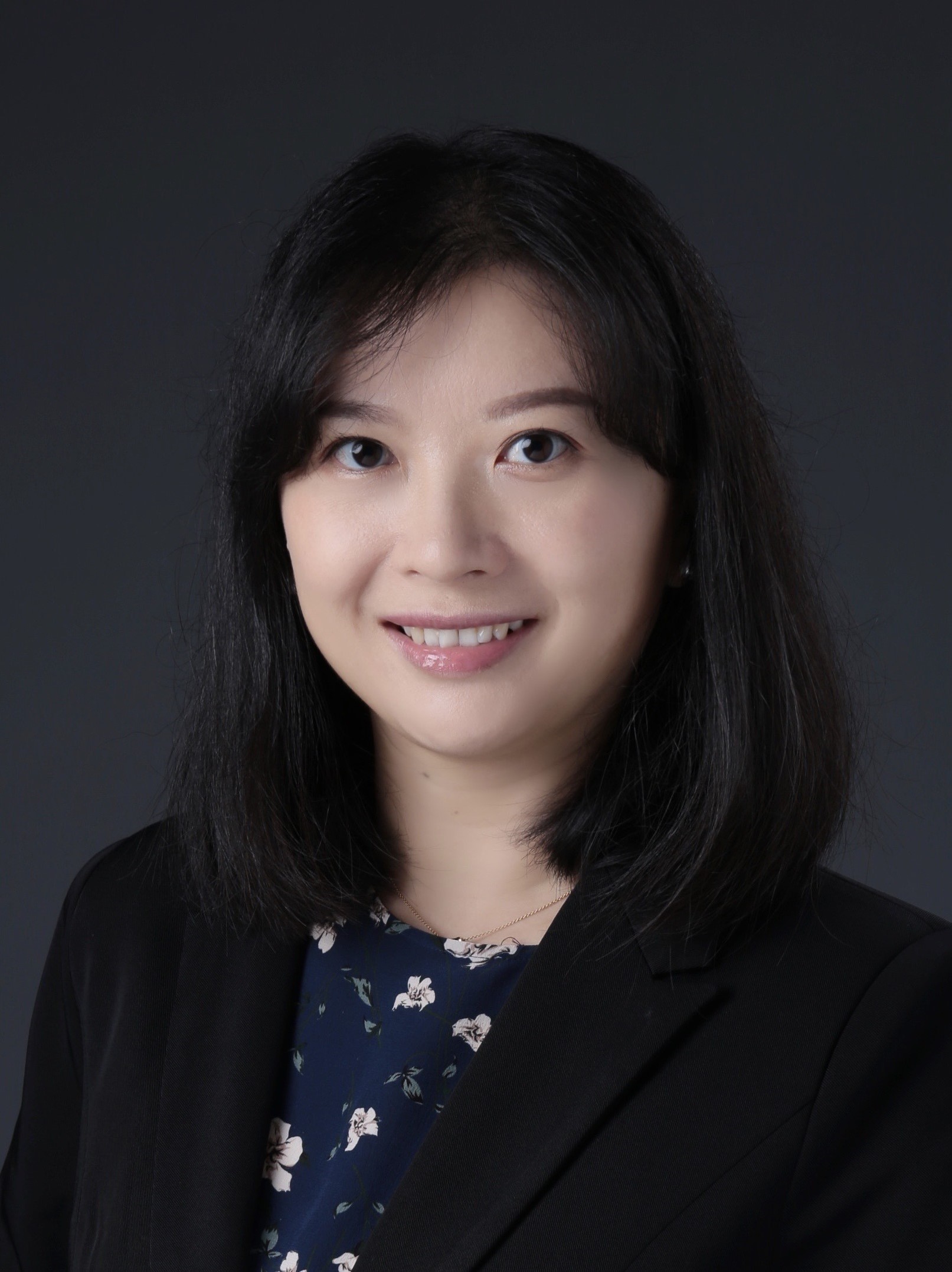}}]{Ya Zhang} is currently a professor at the Cooperative Medianet Innovation Center, Shanghai Jiao Tong University. Her research interest is mainly in machine learning with applications to multimedia
and healthcare. Dr. Zhang holds a Ph.D. degree in Information Sciences and Technology from Pennsylvania State University and a bachelor’s degree from Tsinghua University in China. Before joining Shanghai Jiao Tong University, Dr. Zhang was a research manager at Yahoo! Labs, where she led an R\&D team of researchers with strong backgrounds in data mining and machine learning to improve the web search quality of Yahoo international markets. Prior to joining Yahoo, Dr. Zhang was an assistant professor at the University of Kansas with a research focus on
machine learning applications in bioinformatics and information retrieval. Dr. Zhang has published more than 70 refereed papers in prestigious international conferences and journals, including TPAMI, TIP, TNNLS, ICDM, CVPR, ICCV, ECCV, and ECML. She currently holds 5 US patents and 4 Chinese patents and has 9 pending patents in the areas of multimedia analysis. She was appointed the Chief Expert for the ’Research of Key Technologies and Demonstration for Digital Media Self-organizing’ project under the 863 program by the Ministry of Science and Technology of China. She is a member of IEEE.
\end{IEEEbiography}

\begin{IEEEbiography}[{\includegraphics[width=1in,height=1.25in,clip,keepaspectratio]{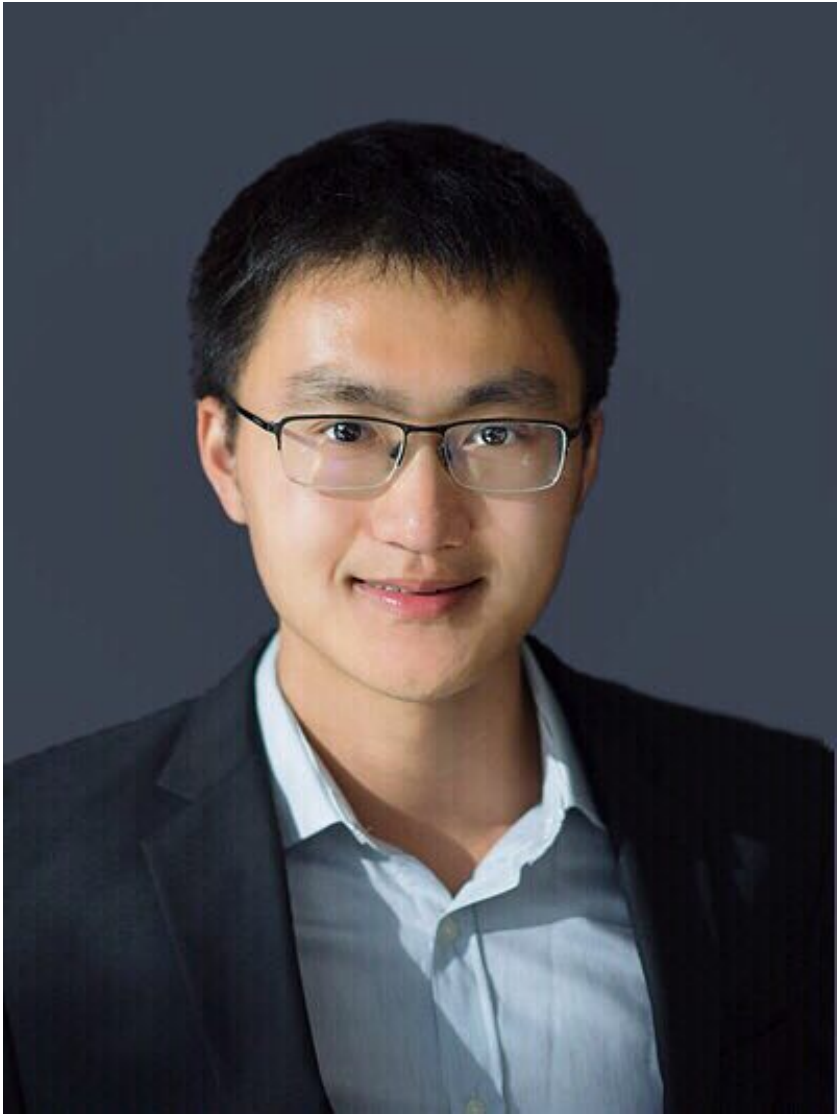}}]{Siheng Chen}  is a tenure-track associate professor of Shanghai Jiao Tong University. Before joining Shanghai Jiao Tong University, he was a research scientist at Mitsubishi Electric Research Laboratories (MERL), and an autonomy engineer at Uber Advanced Technologies Group (ATG), working on the perception and prediction systems of self-driving cars. Before joining industry, Dr. Chen was a postdoctoral research associate at Carnegie Mellon University. Dr. Chen received his doctorate in Electrical and Computer Engineering from Carnegie Mellon University, where he also received two master degrees in Electrical and Computer Engineering (College of Engineering) and Machine Learning (School of Computer Science), respectively. Dr. Chen's work on sampling theory of graph data received the 2018 IEEE Signal Processing Society Young Author Best Paper Award. His co-authored paper on structural health monitoring received ASME SHM/NDE 2020 Best Journal Paper Runner-Up Award and another paper on 3D point cloud processing received the Best Student Paper Award at 2018 IEEE Global Conference on Signal and Information Processing. Dr. Chen contributed to the project of scene-aware interaction, winning MERL President's Award. His research interests include graph neural networks, autonomous driving and collective intelligence. 
\end{IEEEbiography}
\vspace{-1mm}

\appendices

\begin{figure*}[h!]
\begin{center}
\centerline{\subfigure[$\mu_{gt}$]{
         \label{fig:independencea}
 		\includegraphics[scale=0.31]{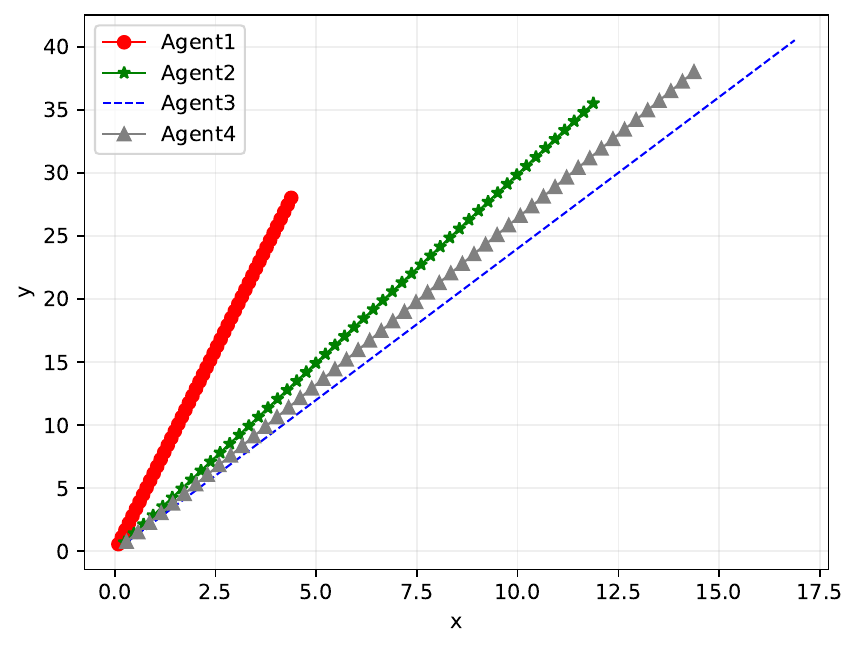}
 	}
 \subfigure[$\epsilon$]{
         \label{fig:independenceb}
 		\includegraphics[scale=0.31]{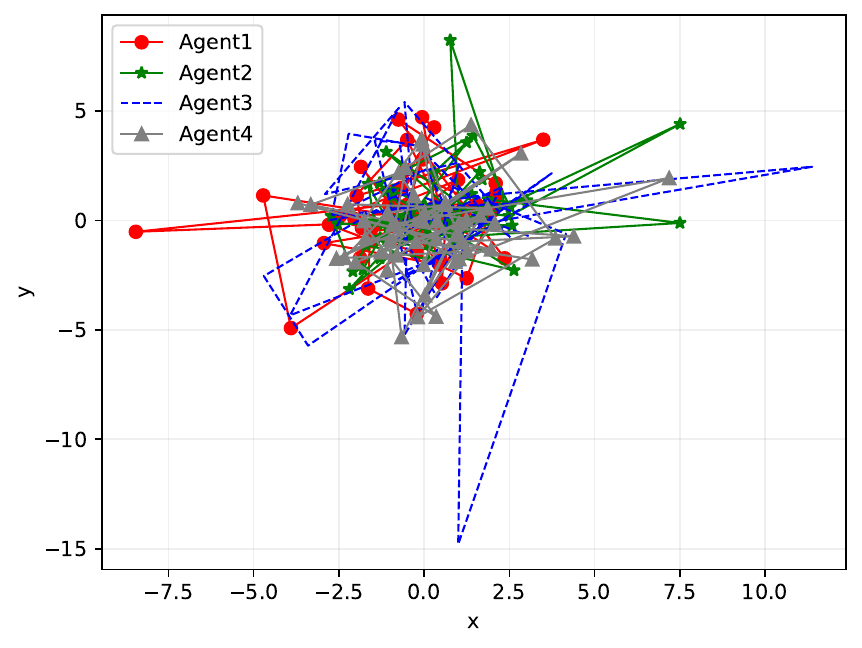}
 	}
 \subfigure[$\x$]{
         \label{fig:independenceb}
		\includegraphics[scale=0.31]{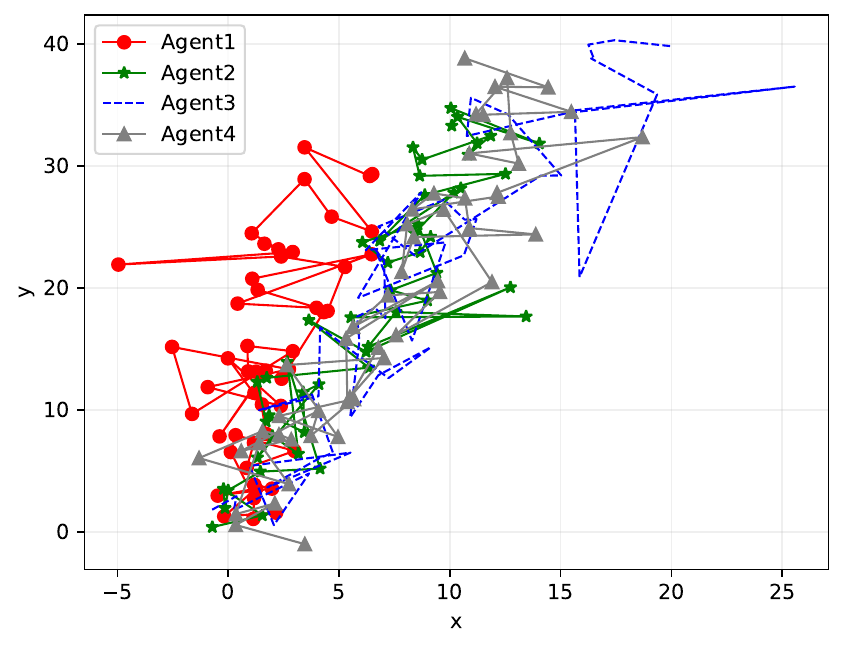}
	}	
	}
\caption{Sample visualization of the generation of Laplace synthetic dataset.}
\label{fig:sy_data}
\end{center}
\end{figure*}

\begin{figure*}[t!]
\begin{center}
\centerline{\includegraphics[width = 2\columnwidth]{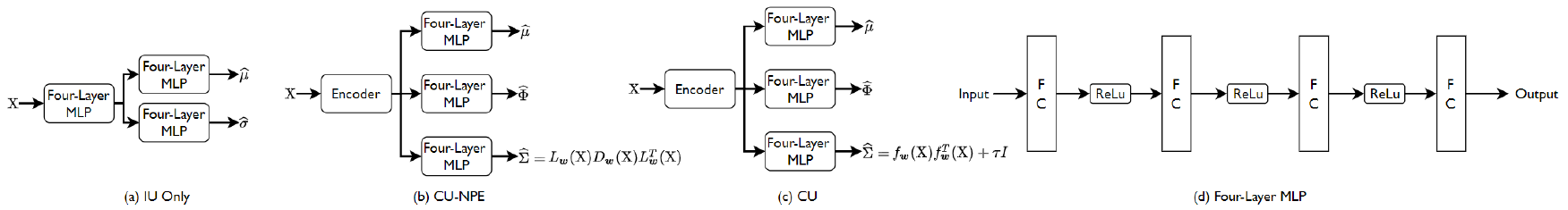}}
\caption{The model structure for IU Only, CU-NPE and CU in the toy problem respectively.}
\label{fig:syn_model}
\end{center}
\end{figure*}

\section{Proof of Laplace Model Design}
\label{ap:PLMD}
\begin{prf}
Consider the $i$-th data sample $(\X^i,\Y^i)$, as in the training process of the prediction model the values of $\X^i$ and $\Y^i$ are given, $p(\Y^i|\pPhi^i,\X^i;\w)$ is a function of $\pPhi^i \in \R^{+}$ with the probability density function: $p(\pPhi^i|\X^i) = \frac{1}{\lambda}e^{-\frac{\pPhi^i}{\lambda}}$:
\begin{align*}
p(\Y^i|\pPhi^i,\X^i;\w)&=f_{\w}(\pPhi^{i}) = \frac{1}{(\pPhi^{i})^{\frac{m}{2}}}e^{-\frac{g_{\w}^{i}}{\pPhi^{i}}},
\end{align*}
where $g_{\w}^{i} = \frac{1}{2}(\Y^{i} - \mu_{\w}(\X^{i}))[\Sigma_{\w}^{-1}(\X^{i})](\Y^{i} - \mu_{\w}(\X^{i}))^{T}$ and $m\in\NN^{+}$ is the number of the agents in the $i$-th data sample. We need to prove that there should exist a $(\pPhi^{i})^{*} \in\R^{+}$ to make:
\begin{align*}
p(\Y^i|\X^i;\w) &=\int_{0}^{+\infty} p(\Y^i|\pPhi^i,\X^i;\w)p(\pPhi^i|\X^i;\w) d\pPhi^i\nonumber\\
&=\int_{0}^{+\infty} f_{\w}(\pPhi^i)p(\pPhi^i|\X^i) d\pPhi^i\nonumber\\
&=E_{\pPhi^i}[f_{\w}(\pPhi^i)]\nonumber\\
&=f_{\w}((\pPhi^{i})^{*})\nonumber\\
&=p(\Y^{i}|(\pPhi^{i})^{*},\X^{i};\w).
\end{align*}
And the existence of $(\pPhi^i)^*$ can be proved by proving a fact that, when $\pPhi^{i}\in\R^{+}$, $f(\pPhi^{i})$ is a continuous bounded function.

As the $g^{i}_{\w}$ can be reformulated as:
\begin{align}
g_{\w}^{i} &= \frac{1}{2}[(\Y^{i} - \mu_{\w}(\X^{i}))L^{\prime}_{\w}(\X)][(\Y^{i} - \mu_{\w}(\X^{i}))L^{\prime}_{\w}(\X)]^{T} \geq 0,
\label{nq}
\end{align}
where $L^{\prime}_{\w}(\X)$ is a lower triangular matrix and the equal sign of (\ref{nq}) is only true when $\mu_{\w}(\X^{i})$ is equal to $\Y^{i}$, but in practice, $\mu_{\w}(\X^{i})$ is hardly equal to $\Y^{i}$, which means in the training process we have:
\begin{align}
g_{\w}^{i}~\textgreater~0.
\label{g_greater}
\end{align}

Based on (\ref{g_greater}), let $\s^i = \frac{1}{\pPhi^i}$, then as $\pPhi^{i}\to0^{+}$, we have $\s^{i}\to+\infty$, so for $\pPhi^{i}\to0^{+}$:
\begin{align}
\lim\limits_{\pPhi^{i}\to0^{+}}f_{\w}(\pPhi^{i}) &=  \lim\limits_{\s^{i}\to+\infty}(\s^i)^{\frac{m}{2}} e^{-\s^i g_{\w}^{i}}\nonumber\\
&= \lim\limits_{\s^{i}\to+\infty}\frac{(\frac{m}{2})!}{(g_{\w}^{i})^{\frac{m}{2}}e^{\s^i g_{\w}^{i}}}=0
\label{lim2}
\end{align}

For $\pPhi^{i}\to +\infty$:
\begin{equation}
\lim\limits_{\pPhi^{i}\to+\infty}f_{\w}(\pPhi^{i}) = 0
\label{lim}
\end{equation}

Furthermore, as the derivative of $f_{\w}(\pPhi^{i})$ is then:
\begin{equation}
\setlength{\abovedisplayskip}{4pt}
\setlength{\belowdisplayskip}{4pt}
f^{\prime}_{\w}(\pPhi^{i}) = (\pPhi^i)^{-\frac{m}{2}-2}\cdot e^{-(\pPhi^i)^{-1} g_{\w}^{i}}\cdot(g_{\w}^{i}-\frac{m}{2}\pPhi^{i}).
\label{de}
\end{equation}
According to (\ref{g_greater}), (\ref{lim}), (\ref{lim2}) and (\ref{de}), when we set $f^{\prime}_{\w}(\pPhi^{i}) = 0$, we can get the maximum value of $f_{\w}(\pPhi^{i})$ is  $f_{\w}(\frac{2g_{\w}^{i}}{m})\in\R^+$.

On the basis of above discussions, when $\pPhi^{i}\in\R^{+}$, $f(\pPhi^{i})$ is a continuous bounded function, which means the $(\pPhi^{i})^{*}$ is existent.
\end{prf}

\section{Toy Problem}

\subsection{Generation Details of Synthetic Datasets}
\label{ap:TP-GDSD}
For the Laplace synthetic dataset, since a random variable $\x$ that obeys a multivariate Laplace distribution $\LL(\mu,\Sigma,\lambda)$ can be formulated as the sum of the mean $\mu$ and $\epsilon$ the product of a Gaussian random variable $g$ and an exponential random variable $\pPhi$: $\x = \mu + g\sqrt{\pPhi}$, where $g \sim \N(0,\Sigma)$ and $\pPhi \sim p(\pPhi) = \frac{1}{\lambda}e^{-\frac{\pPhi}{\lambda}}$ . For generating the Laplace synthetic dataset, we firstly generate 50 different two-dimensional coordinates of four agents that move in a uniform straight line. We denote this part of the data as $\mu_{gt}$, and set it as the mean of the multivariate Laplace distribution to which the trajectories belong. Subsequently, we sample a set of data $\x$ from a multivariate Gaussian distribution $\N(0,\Sigma_{gt})$ (where $\Sigma_{gt} \in \R^{4\times4}$), obviously this set of data contains the information of the covariance matrix of its distribution. Moreover, we sample a set of $\pPhi$ from $p(\pPhi) = \frac{1}{\lambda_{gt}}e^{-\frac{\pPhi}{\lambda_{gt}}}$. We set $g\sqrt{\pPhi}$ as $\epsilon$. Finally, we add the data $\mu_{gt}$ representing the mean value of the distribution and the data $\epsilon$ representing the covariance matrix and $\lambda$ information of the distribution to get our final data $\x$, which is $\x = \mu_{gt} + \epsilon$. At this time, the data $\x$ we get is equivalent to the data sampled from the multivariate Laplace distribution $\LL(\mu_{gt},\Sigma_{gt},\lambda_{gt})$. See Fig.~\ref{fig:sy_data} for an example.
\subsection{Model Structure}
\label{subsec:syn_model}
Fig.~\ref{fig:syn_model} show the model structure of IU Only, CU-NPE and CU used in our toy problem respectively.

\subsection{Metric Computation Details}
\label{ap:TP-MCD}
{$\ell_{2}$ of $\mu$} is the average of pointwise $\ell_2$ distances between the estimated mean and the ground truth mean. {$\ell_1$ of $\Sigma$} is the average of pointwise $\ell_1$ distances between the estimated covariance matrix and the ground truth covariance.

KL is the KL divergence between the ground truth distribution and the estimated distribution $D_{KL}(p_{g}(\!X\!)||p_{e}(\!X\!))$, where $p_{e}(\!X\!)\sim\N(\mu_{p_{e}},\Sigma_{p_{e}})$ is the estimated distribution, $p_{g}(\!X\!)\sim\N(\mu_{p_{g}},\Sigma_{p_{g}})$ is the ground truth distribution, $\Sigma_{p_{e}} \in \R^{k\times k}$ and $\Sigma_{p_{g}} \in \R^{k\times k}$ . For multivariate Laplace distribution, we compute it by the following formula~(\ref{kl_gau}):
\begin{align}
\setlength{\abovedisplayskip}{4pt}
\setlength{\belowdisplayskip}{4pt}
D_{KL}(p_{g}(\!X\!)||p_{e}(\!X\!)) &=p_{g}(\!X\!)\int_{X}[\log(p_{g}(\!X\!)) - \log(p_{e}(\!X\!))]dX\nonumber\\
&= \frac{1}{2}[\log(\frac{|\Sigma_{p_{e}}|}{|\Sigma_{p_{g}}|})\!-\!k\!\nonumber\\
&\quad\quad+(\mu_{p_{g}}\!-\!\mu_{p_{e}})^{T}\Sigma_{p_{e}}^{-1}(\mu_{p_{g}}\!-\!\mu_{p_{e}})\nonumber\\
&\quad\quad+trace\{\Sigma_{p_{e}}^{-1}\Sigma_{p_{g}}\}].
\label{kl_gau}
\end{align}
For multivariate Laplace distribution, as the probability density function of it is too complicated, when we compute the KL divergence, we firstly compute the value of $p_{g}(\!X\!)$ and $p_{e}(\!X\!)$ for each given data sample $X$ respectively, and then we compute $D_{KL}(p_{g}(\!X\!)||p_{e}(\!X\!))$ by the following formula~(\ref{kl_lap}):
\begin{align}
D_{KL}(p_{g}(\!X\!)||p_{e}(\!X\!)) &=\sum_{X}p_{g}(\!X\!)[\log(p_{g}(\!X\!)) - \log(p_{e}(\!X\!))].
\label{kl_lap}
\end{align}


\end{document}